\documentclass[acmsmall,screen]{acmart}

\AtBeginDocument{%
  \providecommand\BibTeX{{%
    \normalfont B\kern-0.5em{\scshape i\kern-0.25em b}\kern-0.8em\TeX}}}

\setcopyright{acmlicensed}
\copyrightyear{2026}
\acmYear{2026}
\acmDOI{XXXXXXX.XXXXXXX}

\usepackage[acronym,nogroupskip,nonumberlist,nopostdot,nohypertypes={acronym,notation}]{glossaries}
\usepackage{sg-macros}

\begin{document}

\title{Neural Radiance Fields for the Real World: A Survey}


\author{Wenhui Xiao}
\email{wenhui.xiao@hdr.qut.edu.au}
\orcid{0000-0002-1933-0451}
\author{Remi Chierchia}
\email{remi.chierchia@hdr.qut.edu.au}
\orcid{0000-0002-0169-6973}
\author{Rodrigo Santa Cruz}
\email{r2.santacruz@qut.edu.au}
\orcid{0000-0002-5273-7296}
\affiliation{%
  \institution{Queensland University of Technology}
  \city{Brisbane}
  \country{Australia}
}
\affiliation{%
  \institution{CSIRO Technology}
  \city{Brisbane}
  \country{Australia}
}

\author{Xuesong Li}
\email{xuesong.li@csiro.au}
\orcid{0000-0001-9295-7333}
\affiliation{%
  \institution{CSIRO Agriculture \& Food}
  \city{Canberra}
  \country{Australia}  
}
\affiliation{%
  \institution{Australian National University}
  \city{Canberra}
  \country{Australia}  
}

\author{David Ahmedt-Aristizabal}
\email{david.ahmedtaristizabal@data61.csiro.au}
\orcid{0000-0003-1598-4930}
\affiliation{%
  \institution{CSIRO Technology}
  \city{Canberra}
  \country{Australia}  
}

\author{Olivier Salvado}
\email{olivier.salvado@qut.edu.au}
\orcid{0000-0002-2720-8739}
\author{Clinton Fookes}
\email{c.fookes@qut.edu.au}
\orcid{0000-0002-8515-6324}
\author{Leo Lebrat}
\email{leo.lebrat@qut.edu.au}
\orcid{0000-0003-1427-4523}
\affiliation{%
  \institution{Queensland University of Technology}
  \city{Brisbane}
  \country{Australia}
}

\renewcommand{\shortauthors}{Xiao, et al.}

\begin{abstract}

Neural Radiance Fields (NeRFs) have remodeled 3D scene representation since release. NeRFs can effectively reconstruct complex 3D scenes from 2D images, advancing different fields and applications such as scene understanding, 3D content generation, and robotics. Despite significant research progress, a thorough review of recent innovations, applications, and challenges is lacking. This survey compiles key theoretical advancements and alternative scene representations and investigates emerging challenges. It further explores applications on reconstruction, highlights NeRFs' impact on computer vision and robotics, and reviews essential datasets and toolkits. By identifying gaps in the literature, this survey discusses open challenges and offers directions for future research.

\end{abstract}



\received{08 January 2025}
\received[revised]{12 November 2025}
\received[revised]{30 June, 2026}
\received[accepted]{13 July 2026}

\maketitle

\definecolor{OK}{HTML}{70e000}
\definecolor{NOPE}{HTML}{EE1C25}
\newcommand{\wendCheck}{\textcolor{OK}{\Checkmark}}
\newcommand{\nope}{\textcolor{NOPE}{\XSolidBrush}}
\newcommand{\add}[1]{\textcolor{red}{#1}}
\newcommand{\del}[1]{\textcolor{red}{\sout{#1}}}

\newacronym{snr}{SNR}{signal-to-noise ratio}
\newacronym{sinr}{SINR}{signal-to-interference-plus-noise ratio}
\newacronym{sir}{SIR}{signal-to-interference ratio}
\newacronym{sqr}{SQR}{signal-to-quantization-noise ratio}
\newacronym{sqnr}{SQNR}{signal-to-quantization-plus-noise ratio}

\newacronym{lvm}{LVM}{large vision model}
\newacronym{ber}{BER}{bit error rate}
\newacronym{evm}{EVM}{error vector magnitude}
\newacronym{isi}{ISI}{intersymbol interference}

\newacronym{bfsk}{BFSK}{binary frequency shift keying}
\newacronym{qam}{QAM}{quadrature amplitude modulation}
\newacronym{mqam}{MQAM}{M-ary quadrature amplitude modulation}
\newacronym{dsss}{DSSS}{direct-sequence spread spectrum}
\newacronym{ofdm}{OFDM}{orthogonal frequency-division multiplexing}
\newacronym{ofdma}{OFDMA}{orthogonal frequency-division multiple access}
\newacronym{fdd}{FDD}{frequency-division duplexing}
\newacronym{tdd}{TDD}{time-division duplexing}
\newacronym{fdma}{FDMA}{frequency-division multiple access}
\newacronym{tdma}{TDMA}{time-division multiple access}
\newacronym{sdma}{SDMA}{space-division multiple access}

\newacronym{ls}{LS}{least-squares}
\newacronym{lms}{LMS}{least mean squares}
\newacronym{omp}{OMP}{orthogonal matching pursuit}

\newacronym{zf}{ZF}{zero-forcing}
\newacronym{mmse}{MMSE}{minimum mean square error}
\newacronym{mse}{MSE}{mean square error}

\newacronym{fft}{FFT}{fast Fourier transform}
\newacronym{dft}{DFT}{discrete Fourier transform}
\newacronym{dtft}{DTFT}{discrete-time Fourier transform}
\newacronym{ctft}{CTFT}{continuous-time Fourier transform}

\newacronym{adc}{ADC}{analog-to-digital converter}
\newacronym{dac}{DAC}{digital-to-analog converter}
\newacronym{fpga}{FPGA}{field-programmable gate array}
\newacronym{enob}{ENOB}{effective number of bits}

\newacronym{rv}{r.v.}{random variable}
\newacronym{svd}{SVD}{singular value decomposition}
\newacronym{sdp}{SDP}{semidefinite programming}
\newacronym{psd}{PSD}{positive semidefinite}
\newacronym{nsd}{NSD}{negative semidefinite}

\newacronym{agc}{AGC}{automatic gain control}
\newacronym{rf}{RF}{radio frequency}
\newacronym{los}{LOS}{line-of-sight}
\newacronym{nlos}{NLOS}{non-line-of-sight}
\newacronym{ple}{PLE}{path loss exponent}
\newacronym[plural=dB,firstplural=decibels (dB)]{db}{dB}{decibel}
\newacronym[plural=dBm,firstplural=decibel milliwatts (dBm)]{dbm}{dBm}{decibel milliwatts}
\newacronym{pa}{PA}{power amplifier}
\newacronym{lna}{LNA}{low noise amplifier}
\newacronym{cw}{CW}{continuous wave}
\newacronym{papr}{PAPR}{peak-to-average power ratio}
\newacronym{tx}{TX}{transmitter}
\newacronym{rx}{RX}{receiver}
\newacronym{sdr}{SDR}{software-defined radio}
\newacronym{usrp}{USRP}{Universal Software Radio Peripheral}
\newacronym{lo}{LO}{local oscillator}
\newacronym{mmwave}{mmWave}{millimeter-wave}
\newacronym{eirp}{EIRP}{effective isotropic radiated power}

\newacronym{csma}{CSMA}{carrier-sense multiple access}
\newacronym{csmaca}{CSMA/CA}{carrier-sense multiple access with collision avoidance}
\newacronym{csmacd}{CSMA/CD}{carrier-sense multiple access with collision detection}
\newacronym{mac}{MAC}{medium access control}
\newacronym{phy}{PHY}{physical layer}

\newacronym{4g}{4G}{fourth generation}
\newacronym{lte}{LTE}{Long-Term Evolution}
\newacronym{5g}{5G}{fifth generation}
\newacronym{nr}{NR}{New Radio}
\newacronym{5gnr}{5G NR}{5G New Radio}
\newacronym{ieee}{IEEE}{Institute of Electrical and Electronics Engineers}
\newacronym{lan}{LAN}{local area network}
\newacronym{wlan}{WLAN}{wireless local area network}
\newacronym{bs}{BS}{base station}
\newacronym{ue}{UE}{user equipment}
\newacronym{ul}{UL}{uplink}
\newacronym{dl}{DL}{downlink}
\newacronym{qos}{QoS}{Quality of Service}
\newacronym{fcc}{FCC}{Federal Communications Commission}
\newacronym{iab}{IAB}{integrated access and backhaul}
\newacronym{hetnet}{HetNet}{heterogeneous network}

\newacronym{siso}{SISO}{single-input single-output}
\newacronym{mimo}{MIMO}{multiple-input multiple-output}
\newacronym{sumimo}{SU-MIMO}{single-user \gls{mimo}}
\newacronym{mumimo}{MU-MIMO}{multi-user \gls{mimo}}
\newacronym{ula}{ULA}{uniform linear array}
\newacronym[\glslongpluralkey={angles of arrival}]{aoa}{AoA}{angle of arrival}
\newacronym[\glslongpluralkey={angles of departure}]{aod}{AoD}{angle of departure}
\newacronym{dof}{DoF}{degrees of freedom}
\newacronym{csi}{CSI}{channel state information}
\newacronym{csit}{CSIT}{\gls{csi} at the transmitter}
\newacronym{csir}{CSIR}{\gls{csi} at the receiver}
\newacronym{cs}{CS}{compressed sensing}

\newacronym{elf}{ELF}{extremely low frequency}
\newacronym{slf}{SLF}{super low frequency}
\newacronym{ulf}{ULF}{ultra low frequency}
\newacronym{vlf}{VLF}{very low frequency}
\newacronym{lf}{LF}{low frequency}
\newacronym{mf}{MF}{medium frequency}
\newacronym{hf}{HF}{high frequency}
\newacronym{vhf}{VHF}{very high frequency}
\newacronym{uhf}{UHF}{ultra high frequency}
\newacronym{shf}{SHF}{super high frequency}
\newacronym{ehf}{EHF}{extremely high frequency}
\newacronym{thf}{THF}{tremendously high frequency}

\newacronym{2d}{2D}{two-dimensional}
\newacronym{3d}{3D}{three-dimensional}

\newacronym{nerf}{NeRF}{Neural Radiance Fields}
\newacronym{inr}{INR}{implicit neural representation}
\newacronym{sdf}{SDF}{Signed Distance Function}
\newacronym{deepsdf}{DeepSDF}{Deep Signed Distance Function}
\newacronym{aabb}{AABB}{axis-aligned bounded box}
\newacronym{3dgs}{3DGS}{3D Gaussian Splatting}

\newacronym{mlp}{MLP}{Multilayer Perceptron}
\newacronym{pe}{PE}{Positional Encoding}
\newacronym{ipe}{IPE}{Integrated Positional Encoding}
\newacronym{fe}{FE}{Feature Encoding}
\newacronym{aafe}{AAFE}{Anti-Aliased Feature Encoding}
\newacronym{she}{SH Encoding}{Spherical Harmonics Encoding}
\newacronym{sh}{SH}{Spherical Harmonics}
\newacronym{sfm}{SfM}{structure-from-motion}
\newacronym{mvs}{MVS}{multi-view stereo}
\newacronym{vit}{ViT}{Vision Transformer}
\newacronym{slam}{SLAM}{Simultaneous Localization and Mapping}
\newacronym{gan}{GAN}{generative adversarial network}
\newacronym{brdf}{BRDF}{Bidirectional Reflectance Distribution function}
\newacronym{ingp}{I-NGP}{Instant Neural Graphics Primitives}
\newacronym{kld}{KL Divergence}{Kullback–Leibler divergence}
\newacronym{ddm}{DDM}{denoising diffusion model}
\newacronym{vae}{VAE}{variational autoencoder}
\newacronym{cnn}{CNN}{convolutional neural network}

\newacronym{fps}{FPS}{farthest point sampling}
\newacronym{fvs}{FVS}{farthest view sampling}
\newacronym{rs}{RS}{random sampling}
\newacronym{hs}{HS}{heuristic sampling}
\newacronym{vmf}{vMF}{von Mises-Fisher}

\newacronym{sota}{SOTA}{State-of-the-Art}
\newacronym{fov}{FOV}{field of view}
\newacronym{ndc}{NDC}{normalized device coordinates}
\newacronym{vr}{VR}{virtual reality}
\newacronym{aigc}{AIGC}{Artificial Intelligence Generated Contents}

\newacronym{cad}{CAD}{Computer-aided design}

\newacronym{crf}{CRF}{camera response function}
\newacronym{hdr}{HDR}{High Dynamic Range}
\newacronym{ldr}{LDR}{Low Dynamic Range}
\newacronym{lr}{LR}{low-resolution}
\newacronym{hr}{HR}{high-resolution}
\newacronym{sr}{SR}{super-resolution}
\newacronym{msi}{MSI}{multi-sphere images}

\newacronym{psnr}{PSNR}{Peak Signal-to-Noise Ratio}
\newacronym{ssim}{SSIM}{Structural Similarity Index Measure}
\newacronym{lpips}{LPIPS}{Learned Perceptual Image Patch Similarity}

\section{Introduction}
\label{sec:intro}

The perception and interpretation of three-dimensional (3D) space fundamentally shapes human interaction with the physical world.
Enabling computer vision systems to understand and process 3D information in the real world is crucial for numerous applications including robotics, tele-health, immersive simulations, and many others.
A key ingredient of a 3D-capable system is the selection of appropriate 3D primitives for representing objects and scenes.
Unlike the \emph{pixel} as a unified representation of 2D images, there are diverse options for 3D representations, e.g. voxels, meshes, and point clouds.
An effective 3D representation must maintain precision and scalability to efficiently represent the 3D world while being learnable to integrate with deep learning frameworks and advanced 3D vision systems.

Traditional 3D representations exhibit significant limitations in satisfying those computational requirements.
While voxels and point clouds are compatible with deep-learning-based frameworks, they suffer from discretization artifacts---voxels face scalability constraints, while point clouds lack connectivity---and yield limited geometry accuracy.
Meshes excel at representing complex geometry; however, their irregular topology and element shapes impede integration with deep-learning frameworks.

Recent research has introduced \glspl{inr}~\cite{park2019deepsdf, mescheder2019occupancy} to overcome these limitations. \Glspl{inr} implement continuous implicit functions by neural networks, enabling accurate and memory-efficient scene representation.
However, the scarcity of high-quality 3D ground-truth data poses a significant challenge for 3D supervision.
To this end, 3D neural rendering has become a central paradigm connecting 2D and 3D. It is grounded in volumetric rendering, which models the physical imaging process where pixel colors are obtained through integration of radiance fields along each ray. The idea of modeling light transport as a radiance function dates back to the 1960s~\cite{chandrasekhar1960radiative} and was formally introduced to computer graphics in the late 1980s~\cite{kajiya1984ray}. 
In the 2000s, subsequent works began to represent 3D scenes directly using color voxel grids~\cite{seitz1999photorealistic,kutulakos2000theory}. In computer vision, differentiable volume rendering began to gain traction for high-quality novel view synthesis.  Early methods incorporated depth cues or stereo-image priors using representations such as soft 3D volumes~\cite{penner2017soft}, Multi-Plane Images (MPI)~\cite{mildenhall2019local}, and voxel-grid decoders~\cite{lombardi2019neural}, but suffered from the scalability limitations. 
Meanwhile, \glspl{mlp} and the PointNet architecture~\cite{qi2017pointnet} have shown a strong ability to learn 3D scene representations from geometric data~\cite{park2019deepsdf,mescheder2019occupancy}. 
Subsequent works~\cite{sitzmann2019scene, niemeyer2020differentiable} addressed the reliance on ground-truth geometry by using image-based rendering losses, but did not account for view-dependent effects.

\textbf{\gls{nerf}}~\cite{Mildenhall2020NeRF}, as known today, arises at the crossroads of these approaches, combining differentiable volume rendering with view-dependent radiance fields to enable photorealistic 3D scene synthesis. 
It introduces an elegant framework to learn a 3D scene as a continuous volumetric function, mapping a 3D position and a view direction to RGB color and density.
Through volume rendering, \gls{nerf} enables the reconstruction of a 3D scene from only 2D images and associated camera poses.
The \gls{nerf} formulation achieves high-quality novel view rendering of a real-world scene while preserving accurate 3D geometry, in contrast to MPI approaches that often introduce specular artifacts. This breakthrough is enabled by several design innovations, including ray batching, differentiable volume rendering, and positional encoding.

Since its introduction in 2020, \gls{nerf} has been drawing significant research attention in both computer graphics and 3D vision communities. 
Fundamental works have enhanced \gls{nerf} models to be more efficient and accurate, addressing challenges posed by complex real-world scenarios, while numerous applications have emerged to advance various 3D vision tasks.
With thousands of up-to-date \gls{nerf}-related research, a comprehensive survey on this topic is both urgent and critical.

\subsection{Our Contributions}
Several early surveys have reviewed the broader landscape of neural rendering~\cite{tewari2022advances} or 2D/3D neural fields~\cite{xie2022neural}, which cover \gls{nerf} as one representative method but consider only a handful of early advancements; conversely, ~\cite{yao2024neural} focuses on progress between 2022 and 2024 and covers only part of the spectrum of \gls{nerf}'s developments.
However, the breadth and rapid progress of \gls{nerf} necessitate an up-to-date, dedicated, in-depth survey.
Although \cite{cai2024nerf} and~\cite{rabby2023beyondpixels} summarize a variety of \gls{nerf}'s improvements, they lack a coherent taxonomy and pay little attention to real-world challenges and applications that are vital for translating \gls{nerf} research into practical deployment.
\cite{zhu2023DeepReviewNeRF} and \cite{gao2022nerf} review \gls{nerf}'s foundational advancements and discuss several applications. Nonetheless, they do not analyze the factors affecting \gls{nerf}'s performance in real-world settings, and omit critical challenges such as uncertainty quantification, dynamic scenes, and indoor environments.
Compared with existing surveys, our work offers the first unified taxonomy that connects \gls{nerf}'s technical developments to real-world deployment challenges and evaluation protocols, as outlined in Table~\ref{tab:comp_survey}.
Given the explosion of \gls{nerf}-related research after 2023, we aim to provide a timely survey of \glspl{nerf}, enabling readers to pinpoint key research directions, match solutions to specific challenges, explore \gls{nerf}'s diverse applications, and gain insights for future work.

To this end, this survey uniquely integrates both foundational and emerging research in a consistent taxonomy.
We distill core progress in improving \gls{nerf}'s rendering quality and efficiency, and clarify how alternative scene representations have evolved from the \gls{nerf} framework.
Beyond methodological advances, we identify factors affecting \gls{nerf}'s robustness and generalizability in real-world scenarios, offering insights into its strengths and limitations.
We further contextualize \gls{nerf} within broader applications, from reconstruction tasks to computer vision and robotics pipelines, highlighting its growing role as a crucial representation across diverse 3D tasks.
To support ongoing research, we curate an extensive collection of publicly available tools, datasets, and benchmarking evaluation protocols. 
To further operationalize this taxonomy, we introduce a complementary synthesis as a decision-oriented comparison of representative method families.
We summarize their representation paradigms, core mechanisms, strengths and limitations, deployment capabilities, and typical application scenarios.
Finally, we identify key research gaps and open challenges, providing perspectives to guide future exploration in \gls{nerf}-related studies.

\begin{table}[!t]\centering
\vspace{-12pt}
\caption{Comparison of our review to other related survey papers. \Checkmark: main focus; $\Delta$: mention.}\label{tab:comp_survey}
\scriptsize
\setlength{\extrarowheight}{3pt}
\resizebox{0.8\linewidth}{!}{
\begin{tabular}{p{0.15\linewidth} |l|c|c|c|c|c|c|c|c|c}\toprule
\multicolumn{3}{l|}{}&~\cite{gao2022nerf} &~\cite{tewari2022advances} &~\cite{xie2022neural} &~\cite{zhu2023DeepReviewNeRF} &~\cite{rabby2023beyondpixels} &~\cite{yao2024neural} &~\cite{cai2024nerf} &Ours \\ \cline{1-11}
\multirow{3}{\linewidth}{Fundamental progress} 
&\multicolumn{2}{c|}{Framework} 
&\Checkmark &\Checkmark &\Checkmark &\Checkmark &\Checkmark &\Checkmark &\Checkmark &\Checkmark\\ \cline{2-11}
&\multicolumn{2}{c|}{NeRF-agnostic enhancements} 
&           &           &           &           &           &           &           &\Checkmark \\ \cline{2-11}
&\multicolumn{2}{c|}{Alternative scene representations} 
&\Checkmark &           &           &           &$\Delta$   &           &           &\Checkmark \\
\cline{1-11}
\multirow{9}{\linewidth}{Real-world challenges} 
&\multicolumn{2}{c|}{Degraded views} 
&\Checkmark &\Checkmark &           &           &           &\Checkmark &           &\Checkmark \\ \cline{2-11}
&\multicolumn{2}{c|}{Sparse training views} 
&\Checkmark &\Checkmark &           &\Checkmark &\Checkmark &\Checkmark &           &\Checkmark \\ \cline{2-11}
&\multicolumn{2}{c|}{Inaccurate camera poses} 
&\Checkmark &\Checkmark &           &\Checkmark &           &\Checkmark &           &\Checkmark \\ \cline{2-11}
&\multicolumn{2}{c|}{Complex light effects} 
&           &\Checkmark &\Checkmark &\Checkmark &           &\Checkmark &           &\Checkmark \\ \cline{2-11}
&\multicolumn{2}{c|}{NeRF-in-the-wild}
&$\Delta$   &$\Delta$   &           &           &           &           &           &\Checkmark \\ \cline{2-11}
&\multirow{3}{*}{Complex scenes}
&Dynamic 
&           &\Checkmark &\Checkmark &            &           &           &           &\Checkmark \\ \cline{3-11}
& &Indoor
&           &           &           &           &           &\Checkmark &           &\Checkmark \\ \cline{3-11}
& &Unbounded
&\Checkmark &$\Delta$   &           &           &           &           &           &\Checkmark \\ \cline{2-11}
&\multicolumn{2}{c|}{Uncertainty quantification} 
&           &           &           &           &           &           &           &\Checkmark \\ \cline{2-11}
&\multicolumn{2}{c|}{Generalizability to unseen scenes} 
&$\Delta$   &\Checkmark &           &           &$\Delta$   &           &           &\Checkmark \\ 
\cline{1-11}
\multirow{4}{\linewidth}{Applications - Reconstruction tasks} 
&\multicolumn{2}{c|}{3D surface} 
&\Checkmark &\Checkmark &\Checkmark &\Checkmark &           &           &           &\Checkmark \\ \cline{2-11}
%
%
&\multicolumn{2}{c|}{Large-scale scenes} 
&\Checkmark &\Checkmark &           &\Checkmark &$\Delta$   &           &           &\Checkmark \\ \cline{2-11}
&\multicolumn{2}{c|}{Medical images} 
&           &           &           &\Checkmark &           &           &           &\Checkmark \\ \cline{2-11}
&\multicolumn{2}{c|}{Digital human} 
&\Checkmark &\Checkmark &\Checkmark &\Checkmark &\Checkmark &\Checkmark &           &\Checkmark \\
\cline{1-11}
\multirow{3}{\linewidth}{Applications - Beyond reconstruction} 
&\multicolumn{2}{c|}{Robotics}
&$\Delta$   &           &\Checkmark &$\Delta$   &           &           &           &\Checkmark \\ \cline{2-11}
&\multicolumn{2}{c|}{3D generation and editing} 
&\Checkmark &\Checkmark &\Checkmark &\Checkmark &\Checkmark &\Checkmark &           &\Checkmark \\ \cline{2-11}
&\multicolumn{2}{c|}{Recognition}
&\Checkmark &           &           &\Checkmark &           &\Checkmark &           &\Checkmark \\ 
\cline{1-11}
\multicolumn{3}{c|}{Datasets} 
&\Checkmark &           &           &           &\Checkmark &\Checkmark &\Checkmark &\Checkmark \\ 
\cline{1-11}
\multicolumn{3}{c|}{Tools} 
&           &           &           &           &           &           &           &\Checkmark \\
\bottomrule
\end{tabular}}
\vspace{-14pt}
\end{table}

\subsection{Scope and Organization}
\begin{figure}[!t]
    \centering
    \begin{tikzpicture}
        \node[anchor=south west, inner sep=0] (image) at (0,0) {\includegraphics[width=1.0\textwidth]{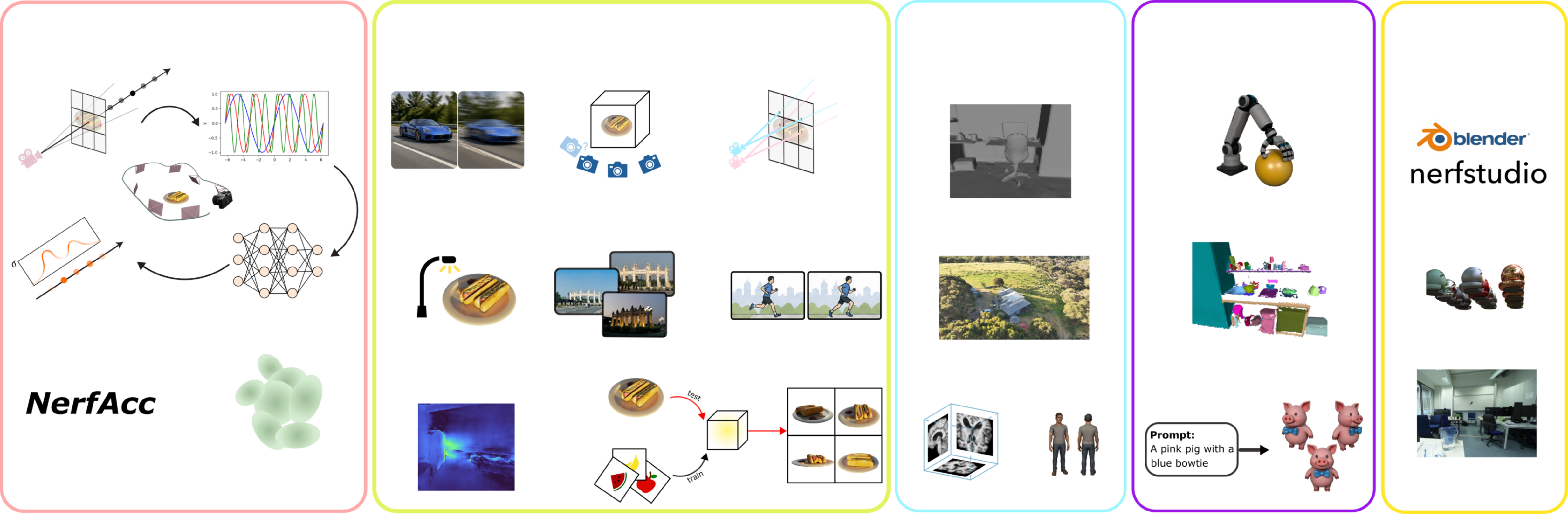}};
        \begin{scope}[x={(image.south east)}, y={(image.north west)}]
            \node[font=\scriptsize, scale=0.9] at (0.12, 0.95) {\textsf{\textbf{Fundamentals (Sec.~\ref{sec:fundamental})}}}; 
            \node[font=\scriptsize, scale=0.9] at (0.4, 0.95) {\textsf{\textbf{Real-World Challenges (Sec.~\ref{sec:challenges})}}};
            \node[font=\scriptsize, scale=0.9] at (0.645, 0.955) {\textsf{\textbf{Reconstruction}}};
            \node[font=\scriptsize, scale=0.9] at (0.645, 0.905) {\textsf{\textbf{(Sec.~\ref{sec:recon})}}};
            \node[font=\scriptsize, scale=0.9] at (0.8, 0.95) {\textsf{\textbf{Other Applications}}};
            \node[font=\scriptsize, scale=0.9] at (0.8, 0.9) {\textsf{ \textbf{(Sec.~\ref{sec:computer_vision})}}};
            \node[font=\scriptsize, scale=0.9] at (0.94, 0.95) {\textsf{\textbf{Tools \&}}};
            \node[font=\scriptsize, scale=0.9] at (0.94, 0.9) {\textsf{\textbf{Datasets }}};
            \node[font=\scriptsize, scale=0.9] at (0.94, 0.85) {\textsf{\textbf{(Sec.~\ref{sec:practice})}}};

            \node[font=\tiny, scale=0.8] at (0.05, 0.895) {\textsf{Sampling}}; 
            \node[font=\tiny, scale=0.8] at (0.05, 0.855) {\textsf{(Sec~\ref{sec2:sampling})}}; 
            \node[font=\tiny, scale=0.8] at (0.17, 0.895) {\textsf{Encoding}};
            \node[font=\tiny, scale=0.8] at (0.17, 0.855) {\textsf{(Sec~\ref{sec2:encoding})}};
            \node[font=\tiny, scale=0.8] at (0.05, 0.39) {\textsf{Volume Rendering}}; 
            \node[font=\tiny, scale=0.8] at (0.05, 0.35) {\textsf{(Sec~\ref{sec2:differential})}}; 
            \node[font=\tiny, scale=0.8] at (0.18, 0.39) {\textsf{Radiance Fields}};
            \node[font=\tiny, scale=0.8] at (0.18, 0.35) {\textsf{Estimation (Sec~\ref{sec2:radiance_architecture})}};
            \node[font=\tiny, scale=0.8] at (0.06, 0.13) {\textsf{NeRF-Agnostic}};
            \node[font=\tiny, scale=0.8] at (0.06, 0.09) {\textsf{Enhancements}};
            \node[font=\tiny, scale=0.8] at (0.06, 0.05) {\textsf{(Sec~\ref{sec2:plugin})}};
            \node[font=\tiny, scale=0.8] at (0.185, 0.10) {\textsf{Alternative Scene}};
            \node[font=\tiny, scale=0.8] at (0.185, 0.06) {\textsf{Representations}};
            \node[font=\tiny, scale=0.8] at (0.185, 0.025) {\textsf{(Sec~\ref{sec2:variants})}};
            \node[font=\tiny, scale=0.8] at (0.13, 0.23) {\textsf{\textbf{3DGS}}};
            \node[font=\scriptsize] at (0.05, 0.63) {\textsf{\textbf{NeRF}}}; 

            \node[font=\tiny, scale=0.8] at (0.29, 0.88) {\textsf{Degraded Views}}; 
            \node[font=\tiny, scale=0.8] at (0.29, 0.84) {\textsf{(Sec~\ref{sec3:views})}}; 
            \node[font=\tiny, scale=0.8] at (0.39, 0.88) {\textsf{Sparse Views}};
            \node[font=\tiny, scale=0.8] at (0.39, 0.84) {\textsf{(Sec~\ref{sec:few_shot_nerf})}};
            \node[font=\tiny, scale=0.8] at (0.5, 0.88) {\textsf{Inaccurate Camera Poses}}; 
            \node[font=\tiny, scale=0.8] at (0.5, 0.84) {\textsf{(Sec~\ref{sec3:camera})}}; 
            \node[font=\tiny, scale=0.8] at (0.3, 0.57) {\textsf{Complex Light}}; 
            \node[font=\tiny, scale=0.8] at (0.3, 0.53) {\textsf{(Sec~\ref{sec3:illumination})}}; 
            \node[font=\tiny, scale=0.8] at (0.4, 0.57) {\textsf{In-the-wild}};
            \node[font=\tiny, scale=0.8] at (0.4, 0.53) {\textsf{(Sec~\ref{sec:nerf_in_the_wild})}};
            \node[font=\tiny, scale=0.8] at (0.51, 0.57) {\textsf{Complex Scenes}};
            \node[font=\tiny, scale=0.8] at (0.51, 0.53) {\textsf{(Sec~\ref{sec3:scene})}};
            \node[font=\tiny, scale=0.8] at (0.3, 0.29) {\textsf{Uncertainty}}; 
            \node[font=\tiny, scale=0.8] at (0.3, 0.25) {\textsf{(Sec~\ref{sec3:uncertainty})}}; 
            \node[font=\tiny, scale=0.8] at (0.465, 0.31) {\textsf{Generalization to New Scenes}};
            \node[font=\tiny, scale=0.8] at (0.465, 0.26) {\textsf{(Sec~\ref{sec3:generalization})}};

            \node[font=\tiny, scale=0.8] at (0.645, 0.84) {\textsf{3D Surface (Sec~\ref{sec4:3d_recon})}}; 
            \node[font=\tiny, scale=0.8] at (0.61, 0.29) {\textsf{Medical}};
            \node[font=\tiny, scale=0.8] at (0.61, 0.25) {\textsf{(Sec~\ref{sec4:medical_recon})}};
            \node[font=\tiny, scale=0.8] at (0.68, 0.29) {\textsf{Human}};
            \node[font=\tiny, scale=0.8] at (0.68, 0.25) {\textsf{(Sec~\ref{sec4:digital_humans})}};
            \node[font=\tiny, scale=0.8] at (0.645, 0.57) {\textsf{Large-Scale Scenes}};
            \node[font=\tiny, scale=0.8] at (0.645, 0.53) {\textsf{(Sec~\ref{sec4:urban})}};

            \node[font=\tiny, scale=0.8] at (0.8, 0.84) {\textsf{Robotics (Sec~\ref{sec5:robotics})}}; 
            \node[font=\tiny, scale=0.8] at (0.8, 0.57) {\textsf{Recognition (Sec~\ref{sec5:recognition})}};
            \node[font=\tiny, scale=0.8] at (0.8, 0.29) {\textsf{3D Generation \& Editing}};
            \node[font=\tiny, scale=0.8] at (0.8, 0.255) {\textsf{(Sec~\ref{sec5:generative_ai})}};

            \node[font=\tiny, scale=0.8] at (0.94, 0.79) {\textsf{Tools (Sec~\ref{sec6:tools})}}; 
            \node[font=\tiny, scale=0.8] at (0.94, 0.53) {\textsf{Datasets (Sec~\ref{sec6:datasets})}}; 
            \node[font=\tiny, scale=0.6] at (0.94, 0.37) {\textsf{\textbf{NeRF Synthetic}}}; 
            \node[font=\tiny, scale=0.6] at (0.94, 0.09) {\textsf{\textbf{ScanNet}}}; 
        \end{scope}
    \end{tikzpicture}
    \caption{Overview of our paper structure, outlining key sections covering NeRF fundamentals and improvement strategies, real-world challenges with corresponding solutions, diverse application domains, and practical resources. Images are adapted from~\cite{EndoNeRF, dai2017scannet, Mildenhall2020NeRF, blender, cai2022pix2nerf, kerbl23gaussiansplatting}.}
    \label{fig:nerf_overview}

\end{figure}

The scope of this survey is summarized in Fig.~\ref{fig:nerf_overview}. 
Our survey covers \gls{nerf}'s major developments in training and rendering, robustness under real-world conditions, applicability across diverse applications, and supporting resources benefiting both academia and industry. 
Building on its central innovation, \gls{nerf} has given rise to a variety of alternative scene representations, such as \gls{3dgs}~\cite{kerbl23gaussiansplatting} for real-time rendering. 
Considering the foundational role of \gls{nerf} as a unifying framework for neural scene representations and its substantial impacts, this survey primarily focuses on \glspl{nerf}, from which most modern architectural designs and core mechanisms for practical challenges can be systematically understood and compared. 
Despite the rise of \gls{3dgs}, this survey does not cover this rapidly evolving line of work; the methodology of \gls{3dgs} has diverged significantly from \glspl{nerf}. We direct our readers to a dedicated survey on this topic~\cite{fei20243dgsSurvey}.

Our organization is as follows.
Section~\ref{sec:fundamental} provides an overview of \gls{nerf} and an in-depth review of fundamental improvement strategies for rendering quality and efficiency, followed by a brief introduction to alternative scene representations in Section~\ref{sec2:variants}.
Then, Section~\ref{sec:challenges} offers a thorough analysis of key real-world challenges and their corresponding solutions.
We explore various \gls{nerf} application tasks including reconstruction (Section~\ref{sec:recon}), robotics (Section~\ref{sec5:robotics}), recognition (Section~\ref{sec5:recognition}), and 3D generation and editing (Section~\ref{sec5:generative_ai}).
Moreover, Section~\ref{sec:practice} compiles frequently used tools and datasets (Section~\ref{sec:practice}).
Towards a practical use of \gls{nerf}-based techniques, we further provide a conceptual and decision-oriented synthesis of key \gls{nerf} methods in Section~\ref{sec1:taxonomy}.
Finally, we conclude by discussing open challenges and key research directions in Section~\ref{sec:open_questions}.

\section{Fundamentals of NeRF}
\label{sec:fundamental}

\Gls{nerf}~\cite{Mildenhall2020NeRF} defines a 3D scene as a continuous volumetric function. Given a 3D position $\mathbf{p} \in \mathbb{R}^3$ in the world system and a view direction $\mathbf{d} \in \mathbb{R}^2$, it outputs the volume density $\sigma \in \mathbb{R}$ and the emitted radiance $\mathbf{c} \in \mathbb{R}^3$. Here, the volume density represents the probability of a ray terminating at the current position.
$\F_{\theta}$ denotes such a learnable function, formally defined by
\begin{equation} \label{eq:nerf}
    \F_{\theta}:(\pos, \dirs) \rightarrow (\colors, \sigma),
\end{equation}
where $\sigma$ depends only on $\pos$, whereas $\mathbf{c}$ depends on both $\dirs$ and $\pos$, to allow for modeling non-Lambertian effects.
The learning process goes through a differentiable rendering procedure: (a) \emph{ray casting and sampling}, (b) \emph{samples encoding}, (c) \emph{radiance field estimation}, (d) \emph{volume rendering}, and (e) \emph{optimization}.

\noindent\textbf{Ray casting and sampling.}
Given the camera's intrinsic and extrinsic parameters, a ray is associated with each pixel. The union of all rays and pixels is used for the supervision of \glspl{nerf}, and for each training step, a subset of rays is \emph{randomly} selected. 
To avoid unnecessary queries on empty or occluded regions while marching rays, points are sampled using \emph{hierarchical volume sampling}. This process involves stratified sampling for a \emph{coarse} model $\F_c$ predicting a rough density distribution, followed by inverse transform sampling from this distribution for a \emph{fine} model $\F_f$. 

\noindent\textbf{Samples encoding.}
Along a ray, positional encoding is applied to each sampled point, which is associated with a 5D coordinate (three for position and two for view direction).
Each element $u$ of the 5D coordinate is normalized to $[-1, 1]$. 
Positional encoding of $u$ is a $L$-levels frequency embedding, implemented by a $\gamma_L$ function mapping $u$ to a higher dimension space $[-1, 1]^{2L}$.
Formally: 
\begin{equation}\label{eqn:freqEncoding}
    \gamma_L(u) = \left(\sin(2^0\pi u), \cos(2^0\pi u), \dots, \sin(2^{L-1}\pi u), \cos(2^{L-1}\pi u)\right).
\end{equation}

\noindent\textbf{Radiance field estimation.}
The estimation of the radiance field is traditionally achieved through two \glspl{mlp} for density and color. For conciseness, the rest of this section will use the original ~\gls{mlp} formulation proposed in~\cite{Mildenhall2020NeRF}.
This backbone can be replaced by more efficient architectures or sparser formulations (see Section~\ref{sec2:radiance_architecture} and Section~\ref{sec2:variants}).

\noindent\textbf{Volume rendering.}
The volume rendering~\cite{kajiya1984ray} involves a continuous integration process through time-parameterized rays.
\Gls{nerf} performs numerical approximation of these integrals via a quadrature rule.
Given a set of discretized time $\{t_i\}_{i=0}^{N}$ along a ray $\br$, the predicted color $\bc_i$, and the density $\sigma_i$; the target pixel's integrated color between $t_0$ and $t_N$ is approximated by $\hat{\bC}(\br)$ and given by:
\begin{gather}
    \label{eq:volume_rendering_discrete}
    \hat{\bC}(\br) = \sum_{i=0}^{N-1} \mathbf{w}_{i}\bc_{i}, \\ 
    \text{where} \quad \mathbf{w}_{i}=T_{i}(1-e^{-\sigma_i\delta_i}), \quad \delta_i = t_{i+1} - t_{i}, \quad \text{and} \quad  T_i = e^{-\sum_{j=0}^{i-1}\sigma_j\delta_j}, \nonumber
\end{gather}
where $\delta_i$ is the distance between contiguous samples $t_{i}$ and $t_{i+1}$, and $T_i$ is the accumulated transmittance measuring the likelihood of a ray traveling from $t_0$ to $t_{i}$ without being intercepted.

\noindent\textbf{Optimization.}
Using the pixel's color $\bC(\br)$ associated with each ray $\br$, the \gls{nerf} model is trained in a supervised fashion by minimizing the total squared error against the reconstructed color $\hat\bC(\br)$.
For each batch of randomly sampled rays $\mathcal{R}$, the parameters of both the \emph{coarse} and \emph{fine} models are optimized simultaneously:
\begin{equation}
    \label{eq:nerf_loss}
    \mathcal{L} = \sum_{\br \in \mathcal{R}} \left( \left \|\hat{\bC}_c(\br) - \bC(\br) \right \|_2^2 + \left \|\hat{\bC}_f(\br) - \bC(\br)\right \|_2^2\right),
\end{equation}
where the subscripts $c$ and $f$ stand for the \emph{coarse} and \emph{fine} models. 

As discussed in \cite{Mildenhall2020NeRF}, 3D \glspl{inr} mark a major advancement in 3D vision and novel view synthesis. However, \gls{nerf}, the most popular method in this field, has a lengthy runtime even on modern GPU workstations.
Improvements on rendering quality and efficiency to the original \gls{nerf} pipeline include sampling (Section~\ref{sec2:sampling}), encoding (Section~\ref{sec2:encoding}), radiance fields estimation (Section~\ref{sec2:radiance_architecture}), and volume rendering (Section~\ref{sec2:differential}). Section~\ref{sec2:plugin} summarizes backbone-agnostic improvements of rendering efficiency and quality, whereas Section~\ref{sec2:variants} introduces alternative scene representations.

\subsection{Sampling}
\label{sec2:sampling}
The Monte Carlo sampling approximation described in Eq.~\eqref{eq:volume_rendering_discrete} is numerically intensive without prior knowledge of the reconstructed geometry. From an unknown density distribution, sampling must be exhaustive to properly approximate the pixel's color $\hat{\mathbf{C}}$, and improvement of the approximation error is linear with the number of points sampled, making it computationally unaffordable.

Higher sampling efficiency can be achieved by wisely selecting integration segments in the non-empty space of the scene. Towards this end, \emph{hierarchical volume sampling} scheme in \gls{nerf} employs a \emph{coarse} model to estimate a rough density distribution, guiding the sampling of the \emph{fine} model and improving numerical approximation efficiency.
However, in Eq.~\eqref{eq:nerf_loss} the \emph{coarse} model's supervision requires rendering but is not used in the final rendered image.

Two lines of research aim to address limitations in this point sampling scheme: proposal-based and occupancy-based methods.
Proposal-based methods e.g. Mip-NeRF 360~\cite{Barron2022MipNeRF360}, replace the \emph{coarse} model with a small \emph{proposal} model, producing only density instead of both density and color. 
However, with this new formulation, the optimization of the \emph{proposal} has to be revisited. Mip-NeRF 360~\cite{Barron2022MipNeRF360} comes up with an online distillation approach, where the \emph{proposal} model is learned using the interlevel loss. 
This loss aligns the predicted weights histogram between the \emph{proposal} model and the final \gls{nerf} model.
The reduction in the size of the \emph{proposal} model allows for a larger \emph{fine} model ($15\times$), contributing to higher synthesis performance with only a modest increase in computational time ($2\times$).
Different from proposal-based methods, occupancy-based methods manage to efficiently rule out sampled points with low density.
For example, PlenOctrees~\cite{Yu2021PlenOctrees} introduces the use of an auxiliary NeRF-SH network trained with a sparsity prior, which is then converted to a sparse Plenoctree data structure to avoid computations on empty spaces of the scene.
~\citet{Hu2022efficientnerf} analyze the weight and density distribution of \gls{nerf}'s sampled points and introduce valid sampling to the coarse stage and pivotal sampling to the fine stage.
Their proposed sampling strategies allow sampling points only in parts of the scene that contribute the most to the final color, allowing an $80\%$ reduction of training time.

Recently, some researchers have emphasized the significance of pixel and view selection in \gls{nerf} training, and have improved \gls{nerf} training efficiency via \emph{ray-} or \emph{view-level sampling methods}.
For instance, ~\citet{Zhang2023FewerRays} enhance training convergence by focusing on pixels with "\textit{dramatic color change}" and regions with high rendering error.
\citet{xiao2024nerfdirector} investigate the role of view sampling in \glspl{nerf} and benchmark different view selection methods to assess their impact. Their findings highlight that with better-selected samples, a \gls{nerf} obtains higher rendering quality with less computational time or supervision.

\subsection{Encoding}
\label{sec2:encoding}

The standard encoding of Eq.~\eqref{eqn:freqEncoding} treats each frequency signal equally, leading to issues when rendering at different resolutions. High-frequency positional encoding features can become aliased, resulting in artifacts known as ``jaggies''.
Additionally, deep networks directly process the radiance field from the encoded input coordinates, which necessitates large and computationally expensive \glspl{mlp} to represent the scene accurately.
Later research on advanced encoding approaches has evolved towards anti-aliasing ability and high efficiency.
We categorize existing encoding approaches into \emph{integrated positional encoding}, \emph{feature encoding}, \emph{anti-aliased feature encoding}, and \emph{compact feature encoding}.

\subsubsection*{Integrated positional encoding}

\begin{figure}[!t]
    \centering

    \begin{tikzpicture}
        \node[anchor=south west, inner sep=0] (image) at (0,0) {\includegraphics[width=0.99\linewidth]{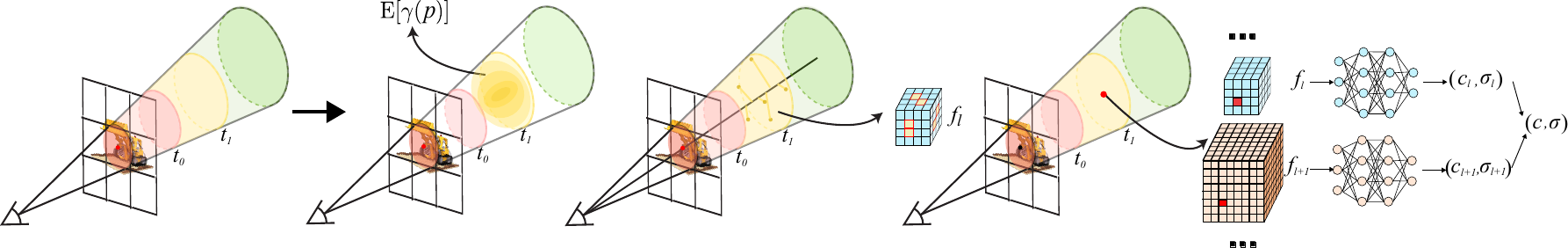}};
        \begin{scope}[x={(image.south east)}, y={(image.north west)}]
            \node[font=\scriptsize] at (0.3, 1.12) {(\textbf{a}) Mip-NeRF~\cite{Barron2021MipNeRF}}; 
            \node[font=\scriptsize] at (0.5, 1.12) {(\textbf{b}) ZipNeRF~\cite{Barron2023ZipNeRF}}; 
            \node[font=\scriptsize] at (0.8, 1.12) {(\textbf{c}) PyNeRF~\cite{Turki2023PyNeRF:Fields}}; 
        \end{scope}
    \end{tikzpicture}

    \caption{Different multi-scale encoding designs.}
    \label{fig:encoding}

\end{figure}

The mipmapping technique~\cite{williams1983pyramidal} is a widely used anti-aliasing approach in the computer graphics rendering pipeline.
To remove high-frequency aliased details in low-resolution signals, one creates mipmaps by downsampling and pre-filtering the texture map to different scales and picking an appropriate scale for rendering.
Inspired by this idea, Mip-NeRF~\cite{Barron2021MipNeRF} extends \gls{nerf} to represent a scene with continuous scales by tracing a cone instead of a ray for pixel rendering, effectively modeling pre-filtered radiance fields.
Mip-NeRF encodes a conical frustum for the \gls{mlp} input via the \emph{\gls{ipe}}.
A conical frustum is approximated for efficient integral computation using a multivariate Gaussian with mean $\mu$ and covariance $\Sigma$, as presented in Fig.~\ref{fig:encoding}(a).
An \gls{ipe} feature $\gamma(\bullet)$ of a volume covered by the frustum can be computed through the expected high-frequency function of the mean and the diagonal of the covariance matrix, which can be mathematically expressed as:
\begin{equation} \label{eq:mipnerf}
    \gamma(\mu, \Sigma) = \mathbb{E}_{\mathbf{p} \sim \mathcal{N}(\mu_\gamma, \Sigma_\gamma)}\left[\gamma(\mathbf{p})\right].
\end{equation}
The cone casting scheme and \gls{ipe} allow Mip-NeRF to encode scale information into features and pre-filter high-frequency aliasing,
ultimately achieving anti-aliasing rendering with multi-scale input views.
However, the computational cost of \gls{ipe} and \gls{mlp}-based decoder remains high.

\subsubsection*{Feature encoding}
The idea behind \textit{\gls{fe}} is to make input features learnable, thereby enabling the training and rendering to be accelerated without using a large \gls{mlp} for inference.
This approach uses auxiliary grid-based data structures, such as voxel octrees~\cite{Sun2021DVGO}, hash grids~\cite{instant-ngp}, or planes~\cite{Chen2022TensoRF}, to store input features.
These features are iteratively updated alongside the model's parameters through gradient backpropagation during training, adapting to the scene characteristics.
Instant NGP~\cite{instant-ngp} is a representative work in this area.
It arranges trainable features into multi-level hash grids, corresponding to different resolutions of voxel grids.
Instant NGP retrieves features from vertices of the voxel $\mathcal{V}_l$ where a 3D point $\mathbf{p}$ is located at each level $l$. 
These multiresolution hash encodings are aggregated using trilinear interpolation to form the level-wise feature $f_{l}$, which is formally expressed as:
\begin{equation}
    f_{l} = trilerp(n_l \cdot \mathbf{p}; \mathcal{V}_l),
    \label{eq:ingp_interpolation}
\end{equation}
where $n_l$ is the grid's linear size.
A compact \gls{mlp} for rendering takes encodings concatenated from each level as its input.
The design of multi-resolution hash grids and the use of small \glspl{mlp} successfully achieves fast training and high-quality rendering with a compact memory size.
Yet, \gls{fe} has limited capabilities to handle scenes with scale changes.

\subsubsection*{Anti-aliased feature encoding}
A recent trend in encoding is \emph{\gls{aafe}}, which combines the advantages of both \gls{ipe} and \gls{fe} for efficient, anti-aliased rendering.
\Gls{ipe} struggles with efficiency because of its reliance on a heavy \gls{mlp} for decoding scale information, while grid-based \Gls{fe} suffers from aliased artifacts.
The challenge of achieving efficient anti-aliased rendering lies in making grid-based data structures scale-aware.
To tackle this, ZipNeRF~\cite{Barron2023ZipNeRF} adapts the complex multi-sample conical frustum used in Mip-NeRF to a multi-resolution hash-grid-based encoding. 
As shown in Fig.~\ref{fig:encoding}(b), ZipNeRF \emph{multisamples} a conical frustum with a set of isotropic Gaussians to encode scale information, querying features through trilinear interpolation across grid levels. This scale-aware interpolation reduces artifacts by \emph{downweighting} high-frequency signals.
ZipNeRF achieves \gls{sota} performance in novel view synthesis.
Despite its improved efficiency over Mip-NeRF~\cite{Barron2021MipNeRF}, ZipNeRF is still slower than \gls{fe}-based architectures like Instant NGP~\cite{instant-ngp} in training time.

Other approaches, such as Mip-VoG~\cite{Hu2023MultiscaleRendering} and Tri-MipRF~\cite{Hu2023trimiprf}, use grid-based structures across scales to predict color and density with scale-aware feature mipmaps.
PyNeRF~\cite{Turki2023PyNeRF:Fields} trains a pyramid of \gls{nerf} models at different grid resolutions, accessing the closest resolutions corresponding to the sample. It then interpolates their scale-aware outputs to produce anti-aliased results compatible with any grid-based \gls{nerf}, as depicted in Fig.~\ref{fig:encoding}(c).
These methods exhibit anti-aliasing capabilities, but they incur a resolution-memory trade-off, as most grid-based architectures do.

\subsubsection*{Compact feature encoding}
Considering the extensive memory cost of grid-based feature encodings, a key trend in advancing feature encoding is to provide more compact and accurate representations.
The successful application of quantization and frequency transform in codecs like JPEG2000~\cite{marcellin2000overview} has inspired some works~\cite{Gsai2023BinaryFields, Rho2022MaskedFields} that incorporate these techniques to increase feature sparsity.
Binary Radiance Fields~\cite{Gsai2023BinaryFields} applies binarization operations to the encoding parameters stored in the grid.
Therefore, these parameters can be stored as either $+1$ or $-1$ instead of expensive float numbers in terms of 16-bit or 32-bit type.
Similarly, ~\citet{Rho2022MaskedFields} propose to use Discrete Wavelet Transform (DWT) and a binarization mask to convert high-dimensional features into masked wavelet coefficient grids.
The feature used for color and density estimation is predicted through inverse DWT.
Run-Length Encoding and Huffman Encoding can be applied to further compress wavelet coefficients and mask bitmaps, yielding a more compact feature representation.

\subsection{Radiance Fields Estimation}
\label{sec2:radiance_architecture}

Advances in radiance field estimation have focused on higher rendering quality, faster training and rendering speed, and more compact models.
In this subsection, we will discuss four mainstream techniques to improve efficiency and compactness.

\subsubsection{Pre-caching}
\label{sec2:pre_caching}

Pre-caching (or \emph{baking}) accelerates rendering by pre-computing and tabulating geometry or appearance data into grid-based data structures like voxel grids.
The inherent challenge is storing view-dependent color information with low memory footprints.

Some methods decompose view-dependent appearance into a combination of basis functions.
For example, PlenOctrees~\cite{Yu2021PlenOctrees} uses \emph{\gls{sh}} functions and stores learned density and \gls{sh} coefficients in the leaves of an octree-based grid.
Similarly, FastNeRF~\cite{Garbin2021FastNeRF} factorizes appearance as the inner product of the view-dependent weights and a position-dependent deep radiance map, which can be cached independently.

Other methods store pre-computed partial information of radiance fields and features, using a small \gls{mlp} to predict view-dependent appearance.
SNeRG~\cite{Hedman2021SNeRG} builds a 3D texture atlas composed of the predicted diffuse color, volume density, and specular features, which are stored in voxel grids.
This 3D texture atlas can be further compressed with advanced codecs like JPEG.
During rendering, SNeRG utilizes a lookup table to query accumulated color and feature vectors to then use a small \gls{mlp}, which evaluates the view-dependent component.
Similarly, MERF~\cite{Reiser2023merf} stores data in a low-resolution 3D voxel grid and three high-resolution 2D planes, which are parameterized as multi-resolution hash encoding.
To further improve rendering efficiency and model compactness, MERF adopts a binary 3D occupancy grid to sparsely store non-zero elements and applies quantization to each entry.
DVGO~\cite{Sun2021DVGO} employs two voxel grids---a coarse one for geometry and a fine one for detailed reconstruction, combined with a small \gls{mlp} for refinement.

\subsubsection{Tensor rank decomposition/Factorization}
\label{sec2:low_rank}

Tensor rank decomposition is a technique to factorize complex feature vectors with multiple low-rank tensor components, enhancing training efficiency and reducing memory footprint.
For example, the traditional CANDECOMP/PARAFAC (CP) decomposition~\cite{carroll1970analysis} represents a tensor decomposition as a sum of multiple rank-one vectors outer products.

\citet{Chen2022TensoRF} apply CP decomposition to model a scene as a tensorial radiance field (TensoRF) with a set of rank-one tensor components, outperforming the vanilla \gls{nerf} model.
This approach enables the use of compact \glspl{mlp} or \gls{sh} to compute the density and directional color from local features.
To capture more complex scene elements, they further introduce the vector-matrix (VM) decomposition, factorizing a tensor as the sum of outer products of vectors and matrices.
The matrices describe the geometry and appearance characteristics of a scene for each plane of a 3D coordinate system.
K-Planes~\cite{Fridovich-Keil2023K-Planes:Appearance} extends the tri-plane factorization to multi-plane factorization, allowing for the representation of dynamic scenes.
Not limited to a simple orthogonal transformation basis, 
Dictionary Fields~\cite{chen2023dictionaryfields} enhances representation capability by using a basis field with periodic transformations and a localized coefficient field, enabling greater diversity in transformation bases.

\subsubsection{Divide-and-conquer}
\label{sec2:divide_and_conquer}

To improve the computational efficiency of a \gls{nerf} model, a common strategy is to decompose a complex \gls{nerf} model into multiple smaller specialized units and then merge their rendering outputs.
Leveraging this strategy, KiloNeRF~\cite{Reiser2021kilonerf} divides a large \gls{nerf} model into thousands of independent small \glspl{mlp}, each of which learns a specific subpart of the scene.
During the volume rendering, this grid of tiny \glspl{mlp} is queried along the ray casts.
The training of these small \glspl{nerf} is achieved through distillation.
A large-capacity \gls{nerf} model is first trained as the \emph{teacher} model.
The distillation is then performed by matching the predicted color and density between the \emph{teacher} model and the locally defined smaller student \gls{nerf} model.
A similar approach, DeRF~\cite{Rebain2020DeRF}, proposes to conduct the Voronoi spatial decomposition of the scene.
Then the Painter's algorithm is applied to efficiently draw the rendering of each Voronoi cell and compose the final rendering.
It can effectively improve the rendering speed under the same quality.
The divide-and-conquer mechanism is also borrowed when the complexity of a scene is too large to be represented by one single model, which is common in large-scale scene reconstruction scenarios discussed in Section~\ref{sec4:city_scale}, allowing to represent a large scene through an ensemble of small models.

\subsubsection{Alternative Architectures}
\label{sec2:others}

Some researchers have revisited the architecture selection and employed the advancements of recent large models or generative models to build a \gls{nerf}-like model and enhance rendering performance, e.g. GNT~\cite{varma2022gnt} and GANeRF~\cite{Roessle2023GANeRF}.
Different from these works,~\citet{Fridovich2022Plenoxels} claim that scene learning can be accomplished without heavy \glspl{mlp}. They propose Plenoxels, an \gls{mlp}-less solution that efficiently learns the radiance field through an interpolation process of nearby voxels for opacity and color estimation. 
In this method, a sparse voxel grid stores the density and \gls{sh} coefficients for each voxel.
A trilinear interpolation on pre-cached information from nearby voxels is applied to model the continuous plenoptic function.

\subsection{Volume Rendering}
\label{sec2:differential}

Classical volume rendering~\cite{kajiya1984ray} assumes a 3D space filled with particles that can absorb, emit, and scatter light, integrating radiance along time-parameterized rays.
\Gls{nerf} simplifies this framework by only considering absorption and emission, using the volume rendering equation in Eq.~\eqref{eq:volume_rendering_discrete} as a cornerstone ingredient in its self-supervised training. 
This allows for efficient optimization of a 3D \gls{inr} from 2D images, but it still suffers in terms of scene representation accuracy and efficiency.

Using the quadrature rule, Eq.~\eqref{eq:volume_rendering_discrete} assumes piecewise constant density and color within the interval $[t_{i}, t_{i+1})$.
This assumption can lead to instability in sampling and ray supervision, causing fuzzy surfaces~\cite{Uy2023PLNeRF}. 
PL-NeRF~\cite{Uy2023PLNeRF} provides a more stable volume rendering approximation by replacing a piecewise constant integration rule with a piecewise linear one, where the weights $\mathbf{w}_i$ and transmittance $T_i$ are given by
\begin{gather}
    \label{eq:plnerf}
    \mathbf{w}_{i}=T_{i}\left(1-e^{-\frac{(\sigma_{i+1}+\sigma_i)\delta_i}{2}}\right), \quad \delta_i = t_{i+1} - t_{i}, \quad \text{and} \quad T_i = e^{-\sum_{j=0}^{i-1}\frac{(\sigma_{j+1}+\sigma_j)\delta_j}{2}}.
\end{gather}

Increasing the order of the approximation scheme leads to increased sampled points within each interval, ultimately decreasing rendering efficiency.
Instead of improving point sampling strategies as discussed in Section~\ref{sec2:sampling}, ~\citet{lindell2021AutoInt} alleviate this approximation issue by directly learning the primitive function of density and color to achieve fast and accurate ray integration. 
More specifically, an efficient AutoInt~\cite{lindell2021AutoInt} framework is proposed to replace the numerical approximation of the integral. The derivative of the coordinate-based network models the original signal during training and reuses the parameters to build an integral network for automatic integration.

\subsection{NeRF-Agnostic Enhancements}
\label{sec2:plugin}

Despite fundamental progress, some works serve as \gls{nerf}-agnostic plugins to enhance either the efficiency or rendering quality.
For instance, NerfAcc~\cite{Li2023NerfAcc:NeRFs} provides flexible APIs fusing operations with CUDA kernels to incorporate efficient sampling methods for \glspl{nerf}.
On the other hand, NeRFLix~\cite{Zhou2023NeRFLix} enhances the rendering quality by correcting \gls{nerf}-style degradations like splatted Gaussian noise, ray jittering, and blurs in a post-processing manner.

\subsection{Alternative Scene Representations}
\label{sec2:variants}

\Gls{nerf} has inspired multiple alternative scene representations that extend beyond the original framework, aiming for higher cross-scene generalization, efficiency, and photorealism.
This subsection reviews these approaches, highlighting their distinctions and benefits compared to the vanilla NeRF framework, with a focus on image-based rendering, explicit geometric representations, and differentiable rasterization.

In vanilla NeRF, the color of a pixel in a novel view is estimated by integrating the radiance field along its corresponding ray. Alternatively, inspired by image-based rendering, the color of a pixel can be obtained through warping, resampling, and blending related pixels from other views. Building on these ideas, methods such as IBRNet~\cite{wang21ibrnet} and PixelNeRF~\cite{yu2021pixelnerf} predict color and density from image features of source views, enabling novel scene rendering without retraining. Subsequent methods, like MVSNeRF~\cite{chen2021mvsnerf} and NeX~\cite{suhail2022light}, further refine multi-view feature aggregation using plane-swept cost volumes and epipolar relationships. These approaches outperform traditional NeRF in generating new views of unseen scenes, and with scene-specific fine-tuning, they can achieve results comparable to \gls{sota} single-scene NeRF.

Another issue with NeRF is the inefficiency in computing the volumetric integral, as evaluating an MLP at numerous sampled points within often sparse 3D space results in long training and rendering times. To alleviate this problem, techniques such as hierarchical sampling~\cite{Mildenhall2020NeRF}, ``baking''~\cite{Reiser2021kilonerf,Chen2022MobileNeRF:Architectures}, and tensor factorization~\cite{Chen2022TensoRF,cao2023hexplane} can be employed. 
Alternatively, some approaches offer a different solution by using explicit geometric structures to support radiance inference. For instance, Point-NeRF~\cite{Xu2022PointNeRF} employs a point-cloud representation along with multi-view features to guide the sampling and computation of radiance and density during ray marching. Similar approaches, like NeuMesh~\cite{yang2022neumesh} and NSVF~\cite{Liu2020NSVF}, leverage triangular meshes and sparse voxel grids, respectively. These methods provide lower-cost inference than NeRF~\cite{Mildenhall2020NeRF} and explicit 3D representations of the geometry, which are key for applications beyond novel view synthesis.

Besides ray marching, neural rendering can also leverage differentiable rasterization methods~\cite{liu2019soft,Laine2020diffrast}, which project 3D primitives directly onto the 2D image plane for rendering.
These methods often exploit explicit primitives that are differentiable and GPU-friendly, such as marching tetrahedrons~\cite{nvdiffrec}, textured polygons~\cite{Chen2022MobileNeRF:Architectures}, and view-lifted point clouds~\cite{kopanas2021point}, and take advantage of optimized and hardware-accelerated rasterization procedures to enable real-time throughput.
A notable method in this domain is \gls{3dgs}~\cite{kerbl23gaussiansplatting}, where scenes are represented as collections of 3D Gaussians, and novel views are rendered by projecting each Gaussian onto the image plane and compositing them via alpha-blending. 
Due to its real-time rendering performance, \gls{3dgs} has become another representative radiance-fields-based method.

\noindent \textbf{Comparison between \gls{nerf} and \gls{3dgs}.}
\Gls{nerf} and \gls{3dgs} differ fundamentally.
\Gls{nerf} is an implicit representation parameterized by neural networks. The ray tracing pipeline enables high-quality geometric learning and accurate modeling of view-dependent appearance; however, rendering remains slow as the network must be queried at hundreds of discrete sample points per ray. 
In contrast, \gls{3dgs} adopts an explicit representation based on anisotropic Gaussian primitives, enabling real-time rendering while maintaining high visual quality. 
Yet, its geometric accuracy and view-dependent appearance modeling may degrade in challenging scenarios, partly due to the ambiguous depth ordering of overlapping Gaussians during rasterization.
Moreover, \gls{3dgs} heavily relies on high-quality \gls{sfm} point clouds for initialization, making its reconstruction quality sensitive to the accuracy and completeness of the underlying point cloud.
Regarding compactness, \gls{nerf} is generally more memory-efficient than \gls{3dgs}, as it encodes a scene using a continuous neural implicit function rather than a large set of explicit geometric primitives concentrated around observed geometry. This also enables flexible modeling within the reconstructed volume.
Furthermore, the implicit formulation makes \gls{nerf} flexible to incorporate additional scene properties (e.g., scene dynamics and semantic information), while extending \gls{3dgs} often requires enriching new attributes to Gaussian primitives or specialized optimization strategies.

\section{Advances for Real-world Challenges}
\label{sec:challenges}

\gls{nerf}-based approaches excel in view synthesis under ideal conditions, such as densely distributed and high-quality training views, accurate camera poses, simple lighting, static objects, and controlled backgrounds.
However, real-world settings often challenge these assumptions, leading to artifacts in synthesized views.
Key challenges include variations in views, camera poses, scene complexity, lighting, uncertainty, and generalizability.
This section illustrates the progress made in addressing these real-world challenges.

\subsection{Degraded Views}
\label{sec3:views}

Traditional \gls{nerf} models are optimized by minimizing the total squared error between predicted and observed pixel values, relying on an accurate correspondence between a ray and its associated pixel value provided by a high-quality training view. 
However, in real-world settings, training views often suffer from different corruptions such as 1) noise and \gls{ldr} in low-light scenes, 2) motion blur and defocus, 3) low resolution, and 4) haze.
These corruptions can result in inconsistent pixel values, leading to poor reconstructions with artifacts when used for training.
This subsection explores the challenges posed to the rendering performance of \glspl{nerf} from the aforementioned degradation types and discusses related advancements.

\subsubsection*{Low-light scenes: noise and low dynamic range}
Images~(\gls{ldr} on default) often fail to represent the \gls{hdr} information of a low-light scene, where there is a high contrast between the brightest and darkest region, lacking high-frequency details.
RawNeRF~\cite{Mildenhall2021RawNeRF} addresses these challenges by optimizing \gls{nerf} directly on RAW data in linear \gls{hdr} color space, applying variable exposure adjustment and tonemapping for better \gls{hdr} signal learning.
Though RawNeRF can produce \gls{hdr} rendering with minimal noise, its requirement of RAW image data has the limitation of large memory footprints.
Other \gls{hdr}-enhancement methods incorporate the physical imaging process upon the \gls{hdr} \gls{nerf} framework to simulate the \gls{ldr} outputs and match the \gls{ldr} input views. 
The simulated imaging process transforms the \gls{hdr} output to the \gls{ldr} one with a tone mapper, which considers exposure information in the image metadata~\cite{Huang2021HDRNeRF} or white balance~\cite{Jun2022HDRPlenoxels}.

\subsubsection*{Blurriness: motion blur and defocus}
Image blurriness from camera motion or defocus cannot be modeled with a static ray, requiring the pixel value to be represented as a weighted sum of sharp rays originating from different spatial positions.
Deblur-NeRF~\cite{ma2022deblurnerf} addresses this by modeling the blurred pixel values as the convolution of the sharp image and a sparse blur kernel. This blur kernel is derived from a canonical sparse kernel and learned view embeddings.
Another approach consists of simulating the physical image formation process throughout the exposure time. Specifically, the rendered blurry image is the weighted sum of a sequence of predicted sharp images captured along the camera's trajectory during this period.
The camera trajectory can be approximated using the Lie-Algebra of $\mathbf{SE}(3)$~\cite{ma2022deblurnerf}, a cubic B-Spline curve~\cite{ma2022deblurnerf}, a Bezier curve~\cite{Lee2023ExBluRF}, or parameterized by a screw axis~\cite{Lee2023DPNeRF}.
DP-NeRF~\cite{Lee2023DPNeRF} further introduces the adaptive combination of sharp rays regarding the relationship between depth and defocus blur.
Additionally, event streams preserve \gls{hdr} details in low-illumination and fast-motion environments and recently have been used for enhancing deblurring and low-light scenes~\cite{
Rudnev2022EventNeRF:Camera, 
Qi2023E2NeRF}.

\subsubsection*{Low resolution}
Learning a \gls{hr} 3D representation of the scene from a set of \gls{lr} images constitutes a \gls{sr} task.
This can be achieved in either \textit{pre-processing}, \textit{in-processing}, or \textit{post-processing} ways.
HRNVS~\cite{Yoon2023HRNVS} works as a \textit{pre-processing} method that super-samples the available training views, providing \gls{hr} samples for training an \gls{hr} \gls{nerf} model.
The \textit{in-processing} \gls{nerf} \gls{sr} method~\cite{Wang2021NeRFSR} simulates the super-sampling scheme by casting multiple rays to each sub-pixel within each pixel.
The learning of each sub-pixel ray is supervised by a sum of weighted errors between each predicted sub-pixel and the corresponding ground-truth integer pixel.
A refinement step is further applied to enhance high-frequency details from an \gls{hr} reference frame.
In \textit{post-processing} approaches~\cite{huang2023refsrnerf, Lin2023FastSRNeRF}, a \gls{sr} module (either single-image-based or reference-based) is used to super-sample the rendered \gls{lr} views.
Patch-based refinement enables the \gls{sr} module to exploit the information embedded in related regions across nearby rays or \gls{hr} reference views.

\subsubsection*{Haze}
Haze in training views introduces the ambiguity between the reconstructed target and light-scattering atmospheric particles, complicating the ray density estimation.
To solve this ill-posed problem, dehazing methods~\cite{ramazzina2023scatternerf, chen2024dehazenerf} incorporate a physics-based scattering model into the volume rendering equation. 
These methods restore a clear scene from a hazy environment by disentangling the actual appearance and geometry of the target from the ambient particles.

\subsection{Sparse Training Views}
\label{sec:few_shot_nerf}
\Gls{nerf} is trained from a collection of posed images. 
Like multi-view stereo, training of \glspl{nerf} intensely relies on fusing information from different viewpoints to successfully extract meaningful 3D information. 
The lack of diversity in view directions provides limited information on the correct geometry, turning the inverse problem of training a \gls{nerf} ill-posed.
This subsection reviews the latest advancements in learning a 3D scene from sparse views or even a single view.

A popular line of work focuses on developing additional geometric constraints for regularizing under-constrained geometry learning.
For instance, \cite{Roessle2022DenseDepthPriors, Wang2023SparseNeRF} leverage depth priors to constrain \gls{nerf}'s training.
Considering external geometric priors are not always available, some research imposes priors based on information theory~\cite{Kim2022InfoNeRF}, frequency bands~\cite{yang2023freenerf}, and density estimation~\cite{seo2023mixnerf}.
Hand-crafted geometric constraints are usually based on specific scenes or architectural hypotheses that cannot be generalized.
Some methods generate training samples in unobserved viewpoints and take advantage of learned priors from pre-trained deep models.
For example, DietNeRF~\cite{Jain_2021_ICCV} generates pseudo labels for both observed and unobserved views, constraining the learned geometry with consistent semantics in a semi-supervised manner.
Rather than solely utilizing hand-crafted or learned priors, DiffusioNeRF~\cite{wynn2023diffusionerf} learns a joint probability distribution of geometry.
The regularization-based methods help reduce visual artifacts in novel view generation. However, they suffer from poor generalization properties, when the queried view is far from the training set.

Concurrent works treat a \gls{nerf} as a decoder, conditioned on either codebooks or image features.
Codebooks-conditioned methods~\cite{Jang2021CodeNeRF, Rebain2022LOLNeRF} retain latent codes for different scene components, e.g., shape and appearance or foreground and background, which condition a \gls{nerf} with a shared latent space in the same category.
Despite category-level view generalizability, these methods are limited to synthetic datasets and struggle with complex backgrounds.
Image feature-based methods~\cite{yu2021pixelnerf, chen2021mvsnerf, hong2024lrm} condition \glspl{nerf} on available 2D image features by projecting 3D points into 2D planes and aggregating pixel-aligned image features, such as PixelNeRF~\cite{yu2021pixelnerf} and MVSNeRF~\cite{chen2021mvsnerf} presented in Section~\ref{sec2:variants}.
ReconFusion~\cite{wu2024reconfusion} further marries diffusion priors with the PixelNeRF-based models for out-painting unobserved geometry and texture.
These image feature-based methods produce good novel view synthesis results from a small number of views or even a single view.
Yet, they are prone to yielding blurry rendering or artifacts in complex real-world settings.

A recent trend is to revisit novel view synthesis as a generative task, aiming to extrapolate the appearance and shape of a scene with limited observations.
To equip \glspl{nerf} with generative ability, this line of work incorporates the \gls{nerf} scene representation into generative models, such as \gls{gan} and diffusion models.
Using the \gls{gan} framework, early approaches build a \gls{nerf} model conditioned on the scene components~\cite{niemeyer2020giraffe} or image features~\cite{chan2022eg3d, gustylenerf}.
Though impressive, these methods may yield inconsistent rendering and lack fine-grained details.
With the great success of diffusion models in 2D image generation, other works manage to leverage diffusion priors to predict unseen details in single-view/sparse-view \gls{nerf} reconstruction. 
They reach this by generating feature encodings with diffusion models~\cite{chen2023ssdnerf, shi2024zerorf}, distilling a 3D scene from the view-conditioned diffusion models~\cite{zhou2023sparsefusion, liu2023zero123}, or employing a \gls{nerf} as the diffusion decoder~\cite{xu2024dmv3d}.
Despite their compelling results, these methods still face the common limitations associated with diffusion models including expensive optimization time, limited rendering resolution, and the inability to handle unbounded scenes.

\subsection{Inaccurate Camera Poses}
\label{sec3:camera}

The training of \glspl{nerf} requires accurate camera poses to precisely learn the 3D scene. Recent works have explored how to estimate or correct pose information during training.
iNeRF~\cite{inerf} is the first to perform pose estimation from a trained \gls{nerf} by optimizing pose parameters to minimize the photogrammetric loss in  Eq.~\eqref{eq:nerf_loss}. This approach can be used to predict the pose of additional images to extend NeRF's training set.
NeRF\textendash\textendash~\cite{wang2022nerfneuralradiancefields} only requires a set of unposed images and allows for computing camera parameters while training a \gls{nerf}. The proposed optimization is robust for translation and rotation noises up to 20\% and $20^\circ$ respectively, limiting its use to forward-facing scenes.
SiNeRF~\cite{Xia2022SiNeRFSN} improves upon \gls{nerf}\textendash\textendash~\cite{wang2022nerfneuralradiancefields} by replacing \gls{nerf}'s activation function with SIREN~\cite{siren} and proposing a mixed region sampling based on SIFT detectors for ray selection instead of random sampling.

Other methods introduce geometric constraints to learn camera parameters more effectively.
BARF~\cite{barf} extends the 2D image alignment problem into 3D by minimizing the photometric loss between pairs of corresponding images through planar alignment. To make their method robust to poor initialization, a coarse-to-fine approach is used by gradually adding higher frequencies in the positional encoding throughout the optimization. 
GARF~\cite{chng2022gaussian} addresses the spectral bias of sine and cosine functions used for positional encoding in Eq.~\eqref{eqn:freqEncoding} (biased towards low-frequency functions). Aimed at this, it uses Gaussian activation functions for a better representation of the first-order derivatives of the encoded signal.
L2G-NeRF~\cite{l2g-nerf} closely follows BARF~\cite{barf} through a new local-to-global approach. 
By combining a flexible pixel-wise alignment and a frame-wise constrained parametric alignment, this method provides improved camera pose computation.

Geometric constraints also provide support for the estimation of camera intrinsics addressing challenges posed by camera distortions and large displacements.
SCNeRF~\cite{jeong2021self} is capable of recovering camera distortion parameters. It introduces a geometric loss called projected ray distance. This reprojection error is computed for 3D correspondences between two views.
NoPe-NeRF~\cite{nope-nerf} focuses on large camera displacements. It demonstrates that correcting monocular depth priors with scale and shift parameters leads to better renderings. It utilizes a monocular depth network to obtain a depth map, from which they extract a point cloud used for a point cloud loss and a surface-based photometric loss. It reports improvements over both BARF and SCNeRF.
Finally, CamP~\cite{camp} employs a camera-preconditioning method to facilitate the optimization process, leading to improved convergence compared to SCNeRF.

Previous methods assume that dense input views are available; with fewer views retrieving accurate camera extrinsics becomes more challenging.
In this context,
SPARF~\cite{sparf} introduces a depth consistency loss, which generates depth maps from training views, to condition the geometry of novel renderings in a pseudo-depth supervision fashion. Additionally, it incorporates a visibility mask estimated from rays’ transmittance to account for occlusions.
SC-NeuS~\cite{sc-neus} focuses on reconstructing accurate geometry by implementing an \gls{sdf} approach, allowing for efficient mapping of 2D features to 3D space. Subsequently, a view-consistent projection loss is used to compute the reprojection error and correct camera poses. Furthermore, it defines a view-consistent patch-warping loss, comparing patches between two views using the homography matrix with a normalized cross-correlation score.

\subsection{Complex Light Effects}
\label{sec3:illumination}
\Glspl{nerf} model expected color as a function of position and view direction but often struggle to capture the appearance of objects under complex real-world lighting. Visual effects such as reflections, occlusions, and shadows also depend on surface characteristics like material properties, smoothness, and environmental lighting. To model these effects, some works have adopted a physics-based approach using the rendering equation~\cite{kajiya1986rendering}. The rendering equation describes the total radiance $L_o(\mathbf{p}, w_o)$ observed at a point $\mathbf{p}$ from a direction $w_o$ as the sum of surface-emitted radiance $L_e(\mathbf{p}, w_o)$ and reflected radiance $L_r(\mathbf{p}, w_o)$. Mathematically,
\begin{equation}
L_o(\mathbf{p}, w_o) = L_e(\mathbf{p}, w_o) + L_r(\mathbf{p}, w_o)
= L_e(\mathbf{p}, w_o) + \int_{\Omega} B(\mathbf{p}, w_i, w_o) L_i(\mathbf{p}, w_i) (n \cdot w_i) \, dw_i,
\end{equation}
where the reflected radiance is the integral over all incoming directions $\Omega$, with incoming radiance $L_i(\mathbf{p}, w_i)$ modulated by a function $B$, the Bidirectional Reflectance Distribution Function (BRDF), and the cosine of the angle between the incident direction $w_i$ and the surface normal $n$.
Further modeling can also account for the light scattering that occurs at the subsurface level. However, this is often neglected in most computer vision applications.

IDR~\cite{idr} and PhySG~\cite{physg} are the first approaches to explore the disentanglement of volumetric representation and appearance in the NeRF context. While IDR focuses on an implicit approach to modeling the scene BRDF, PhySG explicitly estimates light components such as albedo, specular BRDF, and environment map. Both methods jointly learn a surface representation and exploit well-conditioned normals for appearance estimation. However, both assume several approximations and neglect phenomena such as interreflections and occlusions.
%
NeRV~\cite{nerv} estimates a shadow map, surface roughness, and interreflections by approximating environment lighting visibility with a dedicated MLP. However, it assumes known illumination, which limits its applicability. 
NeRD~\cite{nerd}, on the other hand, models illumination using Spherical Gaussians, and implicitly encodes shadows together with the diffuse albedo. 
Ref-NeRF~\cite{ref-nerf} introduces a reflection direction parametrization to simplify the appearance modeling. 
It estimates the color as a linear combination of diffuse and specular components. 
The diffuse component is estimated by the NeRF model, while the specular component results from a directional MLP that encodes surface roughness, surface normal, and view direction.
Ref-NeuS~\cite{ref-neus} builds on Ref-NeRF by estimating an SDF. Additionally, it uses the SDF to access a visibility parameter to refine the reflection score, which mitigates the contribution of a reflective region to the total color loss. 
S3-NeRF~\cite{s3-nerf} exploits shades and shadows to disentangle geometry and appearance. 

An important application in this domain is scene relighting, where new renderings can be generated from different lighting conditions for interactive graphics or animation purposes.
For instance, NeRF-OSR~\cite{nerf-osr} focuses on outdoor scene relighting, closely following NeRD~\cite{nerd}; NeRFactor~\cite{nerfactor} models free-viewpoint relighting assuming a single unknown illumination; and PS-NeRF~\cite{ps-nerf} instead models multiple unknown directional light images.

Other applications explore real-world phenomena.
NeRF-W~\cite{martin2021nerf} tackles novel view rendering for outdoor, unstructured collections of images with different appearances and illumination. NeRFReN~\cite{nerfren} demonstrates that NeRF density has two peaks---one corresponding to the medium depth and the other to the scene reflected or behind glass---and uses this observation to model mirror-like reflections. 
Similarly, works such as ~\cite{zhu2022neural,pan2022sampling,tong2023seeing} focus on rendering scenes through glass, eliminating the undesired reflection and refraction effects. 
Hazy or underwater scenes pose challenges from scattering effects; related methods are detailed in Sec~\ref{sec3:views} and Section~\ref{sec4:underwater}.
Moreover, ORCa~\cite{orca} models the environment's appearance from specular objects on unknown surfaces.

\subsection{NeRF-in-the-Wild} 
\label{sec:nerf_in_the_wild}

\Glspl{nerf} are originally formulated under the assumption that scenes remain geometrically, materially, and photometrically static, such that two observations from an identical camera pose yield identical measurements. Real-world environments rarely satisfy these conditions: illumination changes over time, imaging devices vary in resolution and spectral response, objects move, occlusions appear and disappear, and shadows shift. These factors undermine the baseline assumption of the \gls{nerf} formulation, resulting in rendering artifacts and limiting its applicability beyond controlled laboratory conditions.

The research area extending \glspl{nerf} to unconstrained settings is often referred to as ``NeRF-in-the-Wild''. Within this domain, several approaches tackle different aspects of real-world variability. Photo-tourism~\cite{agarwal2011building} exemplifies the challenge of reconstructing landmarks from unstructured internet photo collections, where even images captured from the same viewpoint on the same day can differ significantly in illumination, shadows, and moving objects. To address these issues, NeRF-W~\cite{martin2021nerf} introduces per-image appearance embeddings to model illumination and camera-dependent variations, alongside transient embeddings to capture and suppress dynamic elements such as moving objects and temporary occlusions. This decomposition improves rendering consistency and mitigates artifacts such as ghosting and color shifts.

These dynamic elements, commonly referred to as \textit{distractors}, have been further explored in subsequent works. RobustNeRF~\cite{sabour2023robustnerf} treats them as outliers within the optimization process, using robust loss functions to suppress their impact. NeRF \textit{On-the-Go}~\cite{ren2024nerf} predicts per-pixel uncertainty and assumes distractors correspond to high-uncertainty regions, which are progressively filtered out during training. IE-NeRF~\cite{WANG25_IENeRF} takes a complementary approach by coupling \gls{nerf} with an inpainting module that generates transient masks to explicitly remove occlusions, thereby refining static scene reconstruction. Together, these strategies improve novel-view synthesis by emphasizing stable scene content.

Beyond robustness to distractors, some \textit{``in-the-wild''} works push NeRF models toward new capabilities. Ha-NeRF~\cite{chen2022hallucinated} hallucinates appearance from single images to synthesize novel views under unseen lighting conditions, eliminating NeRF-W’s per-image optimization requirement. Neural 3D Reconstruction in the Wild~\cite{sun2022neural} extends \gls{nerf} to neural surface reconstruction, enabling geometry recovery resilient to distractors in large-scale photo collections. 
Complementarily, NeRS~\cite{zhang2021ners} introduces a surface-based neural representation that yields closed, reflectance-aware surfaces from sparse, real-world multi-view data, expanding NeRF's utility into surface reconstruction with materials and lighting decomposition. Together, these methods underscore NeRF-in-the-wild’s evolution beyond rendering stability to broader applications in appearance modeling and 3D reconstruction.

Scaling from landmarks to entire urban environments, as in photo-tourism, requires handling massive, heterogeneous image collections captured under inconsistent conditions and frequent occlusions. For example, Urban Radiance Fields~\cite{rematas2022urf} integrate multimodal data for city-scale reconstruction; related large-scale methods are detailed in Section~\ref{sec4:city_scale}.

\subsection{Complex Scene Configurations}
\label{sec3:scene}
The original \gls{nerf} formulation assumes the reconstructed scene to be static, object-centric or front-facing, and spatially bounded.
However, real scenes may violate some of these requirements.
This subsection explores three typical relaxations of these assumptions: dynamic scenes, indoor scenes, and unbounded scenes.

\subsubsection{Dynamic scenes}
\label{sec3:dynamic_scene}

Modeling of dynamic scenes with \gls{nerf} is a complex task that involves designing a neural representation over a four-dimensional (4D) space of 3D space and time. 
A naive idea consists of extending the input with a new dimension for the time component.
However, this often fails to precisely model dynamic scenes and can be very costly to train. Several approaches have been developed to efficiently parameterize neural rendering over time. 

The first group of methods learns a canonical representation of a scene along with a deformation field for each time step. 
Using a time-indexed deformation, novel coordinates for a scene can be computed and substituted directly into the volumetric rendering equation to learn a canonical NeRF representation~\cite{d-nerf, tretschk2020nonrigid}.
Nerfies~\cite{Park2021Nerfies} parameterizes the deformation field on $\mathbf{SE}(3)$. To enforce a robust optimization, penalties for non-rigid transformation and displacement of points in the static background are introduced. The deformations are estimated using a coarse-to-fine regularization scheme. 
~\citet{park2021hypernerf} propose to express \gls{nerf}'s canonical space in a higher dimension, to allow for non-regular deformations such as dynamic topological change.
~\citet{wang2023flow} introduce optical flow constraints for better temporal regularization of video with rapid motion.

Another approach is to learn a coupled representation of appearance and scene flows, capturing the local deformation across neighboring frames. 
In NSFF~\cite{li2020nsff}, the training imposes temporal photometric consistency and cycle consistency between the forward and backward scene flow.
NSFF further considers geometric consistency and single-view depth for added regularization. 
To reduce the need for extensive supervision and enable continuous-time training from a single camera, ~\citet{du2020neural} re-parameterize the flow field with neural ordinary differential equations~\cite{chen2018neural}.  Recently, this formulation has shown \gls{sota} results for complex camera and scene motion using a volumetric image-based rendering framework~\cite{li2022dynibar}.

Compositional approaches separately model fixed and moving scene contents. 
For instance, ~\citet{ost2020neural} learn a scene graph representation to encode independent object transformations and radiance for the static scene. 
However, these frameworks are challenged by complex object deformations and lighting variations. Recent methods further use self-supervised schemes to decompose a dynamic scene into static and dynamic data~\cite{wu2022d2nerf}, or into static, deforming, and newly appearing objects~\cite{song2023nerfplayer}.

Instead of explicitly learning the deformation information, one can parameterize a 4D dynamic scene representation with compact time-conditioned feature vectors. 
N3DV~\cite{li2022n3dv} introduces a \gls{nerf} model conditioned on a set of temporal latent codes, allowing for time interpolation. 
However, this approach suffers from lengthy training time and is challenged by long video sequences. 

Recent methods build an efficient and sparse 4D scene representation by decomposing a dynamic scene into different components, storing them in plane-based~\cite{Fridovich-Keil2023K-Planes:Appearance, song2023nerfplayer} or voxel-based structures~\cite{wang2023mixed, song2023nerfplayer}.
For example, K-Planes~\cite{Fridovich-Keil2023K-Planes:Appearance} decomposes a dynamic scene into six sparse feature planes: three for space representation, and three for spatial-temporal variations.

\subsubsection{Indoor scenes}
\label{sec3:inside_out}

In contrast to traditional \gls{nerf} pipelines, where objects of interest are reconstructed using concentric views, indoor scenes require outward-facing capture, a process often referred to as ``inside-out'' in many NeRF reconstruction codebases.
A typical characteristic of these views is limited overlap between perspectives, with a high prevalence of self-occlusion.
Additionally, they often comprise numerous low-texture regions such as walls and floors, which are well-known obstacles in traditional multiple-view geometry reconstruction.
Without adequate additional supervisory inputs, the novel view synthesis becomes a challenging extrapolation task requiring in-painting of the missing regions.
To tackle these challenges, learning-based priors can be leveraged to provide additional supervision for \glspl{nerf}' optimization.
Some approaches~\cite{Wei2021nerfingmvs, Roessle2022DenseDepthPriors} exploit depth priors to guide the sampling around the actual geometric surface. ~\citet{Roessle2022DenseDepthPriors} also use depth for additional supervision to handle the shape-radiance ambiguity during optimization. 
Semantic information from pre-trained models like CLIP~\cite{radford2021learning} can also provide additional guidance to further constrain \glspl{nerf}' optimization.
It can be utilized as a sampling guide~\cite{qu2023sgnerf}, or a regularizer enforcing cross-view semantics consistency~\cite{kulkarni2023360fusionnerf}.

Notably, the aforementioned methods are designed for perspective views. Another common type of indoor image is the 360$^{\circ}$ panorama.
The scene representation of \gls{msi}~\cite{attal2020matryodshka} has been widely used in 360$^{\circ}$ view synthesis.
Recent approaches~\cite{kulkarni2023360fusionnerf, chen2023panogrf} have adopted a similar design to deal with 360$^{\circ}$ images.
These methods project each 3D position from Cartesian space to spherical space, ensuring the volume rendering is conducted along the actual ray trajectory.

\subsubsection{Unbounded scenes}
\label{sec3:unbounded}
In unbounded scenes, like outdoor 360$^{\circ}$ scenes and street views, objects can be captured from an arbitrary direction and distance.
These scenes are often boundless, posing significant challenges in terms of sampling complexity and model capability when attempting to represent an unbounded learnable space.

\begin{figure}[!t]

    \centering
    \includegraphics[width=0.9\linewidth]{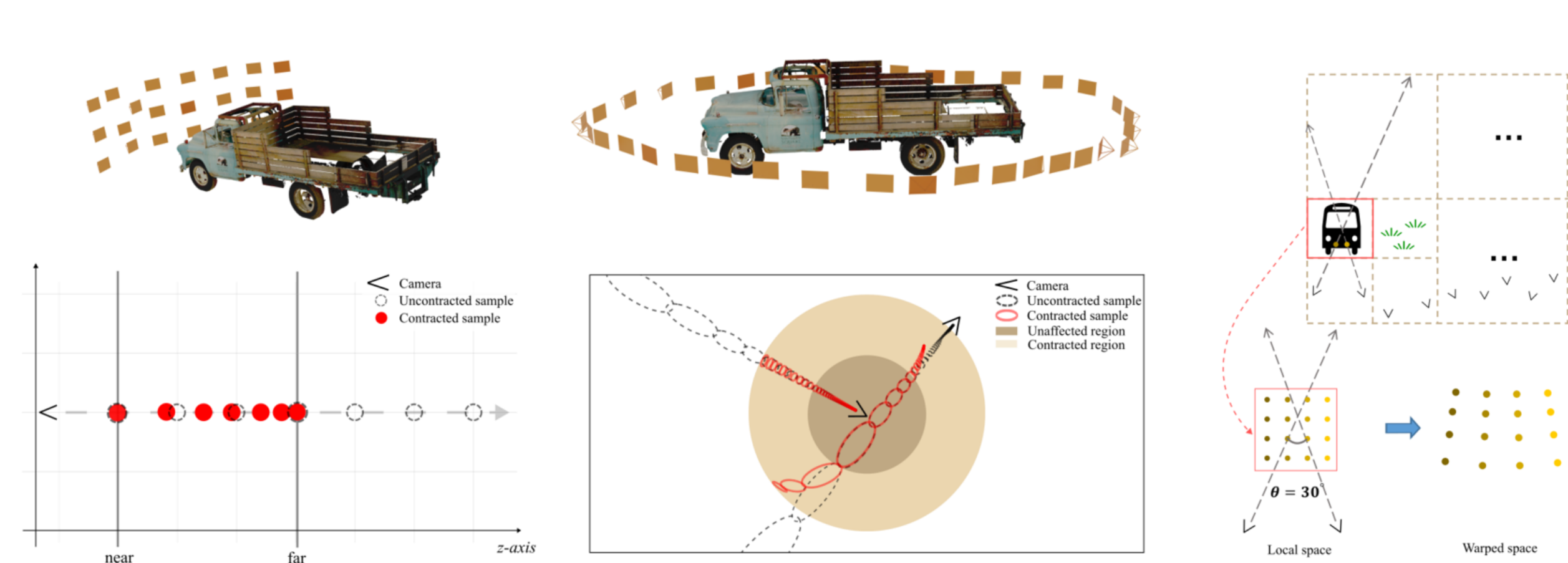}
    \caption{Different contraction functions are applied for different scene settings: (\textbf{left}) a front-facing scene~\cite{Mildenhall2020NeRF}, (\textbf{middle}) a 360$^{\circ}$ scene~\cite{Barron2022MipNeRF360}, and (\textbf{right}) a scene of free camera trajectories~\cite{wang2023f2nerf}. The \textbf{top} row presents the camera trajectories in different settings and the \textbf{bottom} row demonstrates corresponding contraction methods. Figures are adapted from Mip-NeRF 360~\cite{Barron2022MipNeRF360} and F$^2$-NeRF~\cite{wang2023f2nerf}.}
    \label{fig:contraction}

\end{figure}

The original \gls{nerf} addresses these challenges in a front-facing setting by applying the perspective projection that contracts distant samples into \gls{ndc} space along the \textit{z}-axis, as shown in Fig.~\ref{fig:contraction} (left).
Nevertheless, this contraction method is not suitable for unbounded 360$^{\circ}$ scenes where rays aim in all directions.
To parameterize challenging unbounded 360 scenes, an early technical report~\cite{zhang2020nerf++} decomposes the scene into two volumes: an inner sphere volume representing the foreground and an \gls{msi}-based outer volume representing the background.
This design is also adopted in Plenoxels~\cite{Fridovich2022Plenoxels}.
Another line of methods~\cite{Barron2022MipNeRF360, Reiser2023merf, xu2023VRNeRF} aims to parameterize the scene with a novel space contractor function.
For instance, Mip-NeRF 360~\cite{Barron2022MipNeRF360} contracts sampled Gaussians in distant space to a radius-2 ball and correspondingly transforms ray intervals proportional to their disparity, as demonstrated in Fig.~\ref{fig:contraction} (middle).
In practice, cameras can follow an arbitrary trajectory.
To achieve fast scene rendering with free trajectories, F$^2$-NeRF~\cite{wang2023f2nerf} subdivides the scene into regions and introduces a perspective warping to each region, as presented in Fig.~\ref{fig:contraction} (right). This warping ensures that the distances between samples in the warped space align with those observed in visible cameras.

Another challenge in unbounded scene reconstruction is the scale of a scene.
Directly training a \gls{nerf} model on the entire large-scale scene data is impractical due to excessive computational and memory overheads.
Section~\ref{sec4:urban} further explores the challenges and advanced research in applying \glspl{nerf} to large-scale scene reconstruction. 

\subsection{Uncertainty Quantification}
\label{sec3:uncertainty}

Quantifying uncertainty in a \gls{nerf} model is crucial for \glspl{nerf}' practical applications in complex and ambiguous real-world scenes, such as those with limited training data.
The uncertainty in deep learning is typically categorized into two types~\cite{kendall2017uncertainty} --- \emph{aleatoric uncertainty} and \emph{epistemic uncertainty}.
For \glspl{nerf}, \emph{aleatoric uncertainty} refers to the inherent noise (e.g., camera noise) or randomness (e.g., transient objects or illumination changes) in the training data. On the other hand, \emph{epistemic uncertainty} considers the uncertainty within the entire system due to model misspecification or insufficient training data, such as limited available training views.

NeRF-W~\cite{martin2021nerf} paves the way for uncertainty quantification of \gls{nerf} models by reparameterizing the \gls{nerf} model with an additional output of uncertainty.
It quantifies the aleatoric uncertainty and mitigates the influence of transient scene elements.
A similar method is used~\cite{Ran2022NeurAR} to quantify the uncertainty for the next best view planning problem.
Following the Bayesian approach, S-NeRF~\cite{shen2021snerf} and its follow-up CF-NeRF~\cite{Shen2022ConditionalFlowNeRF} assess the epistemic uncertainty based on variational inference by sampling and estimating an approximate posterior distribution across multiple potential radiance fields and optimizing the network.
Nonetheless, the aforementioned uncertainty quantification methods need to reformulate the \gls{nerf} architecture and modify the training pipeline.
It is difficult to introduce these reformulations to more contemporary models like Instant NGP, which comprise complex sampling mechanisms and feature encoding designs.
Instead, Bayes' Rays~\cite{goli2024bayes} does not require modifying the \gls{nerf} training and estimates the posterior distribution through a Laplace approximation based on model perturbations.
Despite compatibility with any \gls{nerf} architecture, it can only assess epistemic uncertainty.
In contrast, Density-aware \gls{nerf} Ensembles~\cite{Sunderhauf2022DensityAwareEnsembles} quantifies both the aleatoric and epistemic uncertainty of the \gls{nerf} as the sum of the variance of predicted color and termination probability across multiple observations.
It does not require modifying \gls{nerf} training.
However, the ensemble-based approach requires multiple substantial computations for one \gls{nerf} model, which has efficiency limitations considering broad real-world applications.

\subsection{Generalizability to Unseen Scenes}
\label{sec3:generalization}

The vanilla \gls{nerf} encodes a 3D scene into the weights of coordinate-based \glspl{mlp}, requiring per-scene optimization.
As such, it still relies on a costly optimization from scratch each time it encounters new scenes and geometries.
A generalizable \gls{nerf} should indeed be capable of novel view synthesis on an unseen scene with minimal retraining or without retraining.
Aiming at this, pioneering research overlaps with partial research within conditional \glspl{nerf} discussed in Section~\ref{sec:few_shot_nerf}.
Codebooks-conditioned methods e.g. LOLNeRF~\cite{Rebain2022LOLNeRF} share category-level 3D prior knowledge with scene latent codes such that it can predict out-of-distribution scenes after fine-tuning the scene latent codes during inference.
Adopting an image encoder to predict the latent code and view direction for \gls{gan}-based \gls{nerf}, Pix2NeRF~\cite{cai2022pix2nerf} can generalize to unseen scenes without requiring inference-time fine-tuning.
However, these methods are dependent on prescribed instances within the same category.
Image feature-based methods like PixelNeRF~\cite{yu2021pixelnerf}, MVSNeRF~\cite{chen2021mvsnerf}, and GRF~\cite{trevithick2021grf} extract general image features with a 2D \gls{cnn}.
They present the potential to generalize to unseen scenes across categories while producing blurry renderings with artifacts.

More recently, a new trend is to leverage recent advances in \glspl{lvm} for enhancing cross-scene generalization performance.
Some approaches obtain a generic 3D prior using vision transformers.
For instance, VisionNeRF~\cite{lin2023visionnerf} uses a 2D \gls{cnn} for local image information and a pre-trained vision transformer~\cite{Dosovitskiy2021ViT} for global information.
LRM~\cite{hong2024lrm} employs a pre-trained vision model to extract image features, which are then projected to tri-planes~\cite{Chen2022TensoRF} using a transformer decoder via a cross-attention mechanism.
Furthermore, GNT~\cite{varma2022gnt} restructures the entire architecture with two transformer-based modules---a view transformer, for aggregating multi-view image features, and a ray transformer, for color predictions, instead of volume rendering.
These methods demonstrate impressive performance in single-view reconstruction and cross-scene generalization.
Nevertheless, they remain challenged when relying solely on semantic information to recover fine-grained details in unobserved regions.

\section{Reconstruction Tasks}
\label{sec:recon}

A typical use case of \gls{nerf} is 3D surface reconstruction (Section~\ref{sec4:3d_recon}), large-scale scenes (Section~\ref{sec4:urban}), medical-related modalities (Section~\ref{sec4:medical_recon}), and digital humans (Section~\ref{sec4:digital_humans}).

\subsection{3D Surface Reconstruction}
\label{sec4:3d_recon}

\glspl{nerf} can generate novel views, but extracting high-quality surfaces from the density-based scene representation using methods such as marching cubes results in poor outcomes due to insufficient constraints on surface~\cite{neus}. To overcome this problem, several methods embed surface representations into \glspl{nerf} formulations, either using implicit or explicit representations.

\subsubsection{Implicit representations}
Implicit surface representations, such as occupancy fields and distance fields, can model many different geometries.
Although UNISURF~\cite{unisurf} manages to obtain accurate surfaces through occupancy fields,
\acrfull{sdf} have become the preferred alternative for surface representations in \gls{nerf} frameworks.

An \gls{sdf} is an implicit representation that describes a geometry through a signed distance function. Such a function can be learned by a \gls{nerf} and is denoted $s_\theta : \mathbb{R}^3 \rightarrow \mathbb{R}$. Using this representation, the geometry of the scene can be easily extracted by computing the 0-level set of $s_\theta$.\footnote{We recall that the $\ell$-level set of a function $f$, is given by
$L_\ell(f) = \left\{ \mathbf{p} \in \mathbb{R}^3 | f(\mathbf{p}) = \ell \right\}$.}
Leveraging this representation, approaches such as NeuS~\cite{neus} and VolSDF~\cite{volsdf} propose different formulations to derive the density $\sigma$ from the \gls{sdf}.

\noindent NeuS~\cite{neus} defines the density function to be maximal at the zero-crossings of the \gls{sdf} $s_{\theta}$ as,
\begin{equation}\label{eq:neus}
 \sigma(\mathbf{p}) = \max \left( \frac{-H_\tau'(s_\theta(\mathbf{p}))}{H_\tau(s_\theta(\mathbf{p}))} , 0\right) \,, \\
 \text{where}~H_\tau(u) = \frac{1}{1+e^{-\tau u}} \,,
\end{equation}
where $\tau$ is a hyper-parameter that controls the sharpness of the logistic sigmoid function $H$. 

\noindent Similarly, VolSDF~\cite{volsdf} defines,
\begin{equation}
    \sigma(\mathbf{p}) = \alpha\Psi_{\beta} (-s_{\theta}(\mathbf{p})), \\
    \text{where}~\Psi_{\beta}(u) = \left\{
    \begin{array}{lc}
         \frac{1}{2}\exp{(\frac{u}{\beta})} & \ \text{if} \ u\leq0 \\
          1-\frac{1}{2}\exp{(-\frac{u}{\beta})} & \ \text{if} \ u>0
    \end{array}
    \right. \,,
\end{equation}\label{eq:volsdf}
where $\alpha$ and $\beta$ are learnable parameters.

Other methods explore variations of the \gls{sdf} to customize surface representation.
3PSDF~\cite{3psdf} introduces the NULL sign to represent non-closed surfaces by excluding surface regions from the optimization pipeline.
NeuralRecon~\cite{neuralrecon} utilizes Truncated Signed Distance Functions (TSDF) to represent local surfaces rather than optimizing the whole geometry at once.
These parametrizations preserve the advantages of traditional \glspl{nerf} while learning smooth surfaces.

Due to their watertight nature, \gls{sdf} formulations struggle to represent complex geometries like closed surfaces with holes or nested surfaces. Unsigned Distance Functions (UDFs) can represent such surfaces by eliminating the sign convention~\cite{ndf}.
However, UDFs suffer from two limitations: they cannot account for occlusions, and their non-negative behavior causes unstable gradient computation on the 0-level set due to cusps, which are non-differentiable~\cite{neuraludf}. 
To overcome these challenges, NeuralUDF~\cite{neuraludf} introduces a differentiable visibility indicator function approximating the surface with a soft probabilistic distribution rather than relying on a hard threshold.

\subsubsection{Explicit representations}
Explicit representations describe surfaces using parametric geometries (e.g. meshes or point clouds). 
NVDIFFREC~\cite{nvdiffrec} employs a coarse-to-fine approach to estimate an \gls{sdf} on a deformable tetrahedral grid. Via differentiable rendering for the scene's appearance, they extend previous work~\cite{dmtet} to fit 2D supervision of \gls{nerf} approaches.
VMesh~\cite{vmesh} uses a hybrid volume-mesh representation merging geometry and volume primitives. Following discretization, they store texture, geometry, and volumetric information in efficient explicit assets.
~\citet{wang2023adaptive} utilize spatially varying kernel sizes $\tau$ (in Eq.~\eqref{eq:neus}) for shell extraction of different frequency details (e.g. smooth surfaces or volumetric content such as clouds). Via dilation and erosion of the explicit shell, they reduce the number of samples for smooth surfaces while maintaining a sufficient amount for high-level details of volumetric content. Although these explicit surface representations are often less flexible than implicit surfaces, they offer faster training and rendering and enable the direct implementation of geometric constraints.

\subsubsection{Surface Losses}
Accessing a surface representation during training allows adding surface regularization terms to the neural rendering loss.
With regard to implicit methods,
~\citet{grop2020implicit} show that it is possible to obtain a smooth and accurate representation by employing the Eikonal term, which encourages the representation to have a unit-norm gradient.
To further encourage smoothness, Neuralangelo~\cite{neuralangelo} regularizes the mean curvature of their \gls{sdf} representation.

Explicit methods, on the other hand, often customize regularization terms to their representation. NVDIFFREC~\cite{nvdiffrec} penalizes the \gls{sdf} sign change at the edges of the tetrahedral grid, reducing random faces and floaters.
VMesh~\cite{vmesh} optimizes the mesh by comparing its silhouette and depth map with the opacity and depth value rendered from the initial \gls{sdf} stage.
~\citet{wang2023adaptive} encourage smoothness of a shell representation by regularizing its spatially varying kernel size.

\subsubsection{Additional Surface Supervision}
To extend the control over surface learning, various methods utilize information computed by the \gls{sfm} step or rely on off-the-shelf methods for additional priors.
Geo-Neus~\cite{geo-neus} and RegSDF~\cite{regsdf} directly supervise their \gls{sdf} using the 3D sparse point cloud generated from \gls{sfm} algorithms (e.g. COLMAP~\cite{schonberger2016structure}). 
Following the assumption that those points lie on the surface, they enforce the \gls{sdf} values to be zero. 
~\citet{darmon2022improving} use additional multi-view constraints and structural similarity to better recover high-frequency details.
When only a few views are available, surface methods significantly deteriorate. To overcome this issue,
s-VolSDF~\cite{s-volsdf} leverages \gls{mvs} priors to guide implicit surfaces from sparse views. In turn, the reconstructed depth information refines MVS depth sampling, providing more accurate estimations.
MonoSDF~\cite{monosdf}, instead, uses monocular geometric priors predicted by a pre-trained network to directly optimize depth and normal estimations. Although these approaches yield more accurate and smoother surfaces, they come at the cost of additional supervisory data or pre-trained models.


\subsection{Large-scale Scene Reconstruction}
\label{sec4:urban}
Large-scale scene reconstruction presents unique challenges for \gls{nerf}, due to the large data volumes that must be stored and processed, the variability in sensing conditions arising from extended capture times in complex environments such as city-scale landscapes, and modifications required in the rendering function to handle varied scenarios like underwater scenes.

\subsubsection{Remote Sensing Scene Reconstruction}
\label{sec4:remote_sensing}

Reconstructing scenes using satellite or aircraft sensing data is challenging due to time, lighting, weather variations, limited viewing angles, and moving objects. 
Stereo-matching approaches~\cite{de2014automatic, perko2019mapping} are commonly used in satellite-based photogrammetry but face difficulties with lighting inconsistencies from shadow movements and sky conditions. 
\Gls{nerf}, with its implicit neural scene representation, offers a novel solution. 
S-NeRF~\cite{derksen2021shadow} models direct sunlight through a local visibility field and learns indirect illumination from diffuse light via a non-local color field. Performance is improved for 3D reconstruction and novel view synthesis, but transient objects remain an issue. 
Sat-NeRF~\cite{mari2022sat} employs a task-uncertainty learning approach~\cite{nerf-w} to model transient objects and uses rational polynomial coefficient camera models~\cite{grodecki2001ikonos} to assist point sampling, yielding better results. 

\subsubsection{Underwater Reconstruction}
\label{sec4:underwater}
The original \gls{nerf}~\cite{Mildenhall2020NeRF} and many of its variants assume that multi-view images are captured in a non-hazy medium, focusing solely on radiance from foreground and background objects. This assumption fails in real-world scenarios like foggy or underwater environments, where scattering effects are present. 
For example, SeaThru-NeRF~\cite{levy2023seathru} directly integrates the scattering medium into its rendering equation through a wavelength-independent attenuation model, using two wideband coefficients to represent the medium.

\subsubsection{City-Scale Reconstruction}
\label{sec4:city_scale}
Unlike the original \gls{nerf}, which performs well in small, controlled settings, city-scale scenes present three key challenges:
various levels of detail arising from large variations in viewing distance, real-world complexities (e.g., moving objects, weather changes, and dynamic lighting), and long rendering times for large image sets. 

City-NeRF~\cite{Bungeenerf} employs a progressive approach, increasing model detail as views get closer, efficiently managing multi-scale complexities in urban environments. 
BlockNeRF~\cite{Tancik2022BlockNeRF} improves scalability by dividing large datasets into manageable blocks, enabling parallel processing and reducing computational demands. This approach is particularly effective for rendering extensive city models.
Switch-NeRF~\cite{zhenxing2022switch} introduces a learnable gating network to dynamically assign 3D points to sub-networks, optimizing scene decomposition and radiance field reconstruction without hand-crafted segmentation.
Mega-NeRF~\cite{Turki2022Mega-NeRF} further optimizes large-scale scene reconstruction and rendering through a scalable architecture, enabling detailed urban fly-throughs for virtual reality and geographic exploration.
UE4-NeRF~\cite{gu2024ue4} partitions large scenes into sub-NeRFs, represented by polygonal meshes, and integrates \gls{nerf} with Unreal Engine 4 (UE4) to achieve fast rendering at 4K resolution with high frame rates (up to $43$ FPS). 
%
%
Finally, Urban neural fields~\cite{rematas2022urf} adapts \gls{nerf} for dynamic urban elements, enabling realistic reconstruction of scenes with moving vehicles and changing lighting.
\subsection{Medical Images Reconstruction}
\label{sec4:medical_recon}

Cameras are used in several medical applications such as skin inspection for wounds, cancer detection, and endoscopic examination.
Similarly to other applications, \gls{nerf}'s ability to generalize to unseen views enables ~\cite{psychogyios2023realistic} to augment the limited endoscopic-imaging datasets.

EndoNeRF~\cite{EndoNeRF} models human anatomy using \glspl{nerf} from surgical video. The authors show how \glspl{nerf} could improve upon prior \gls{slam}-based approaches for estimating non-rigid tissue deformation. They draw inspiration from D-NeRF~\cite{d-nerf} which maps the deformation space to a trained canonical space. EndoNeRF exploits stereo views used in an operating room to estimate coarse depth maps for additional supervision.

Additional constraints using \gls{sdf} are investigated in EndoSurf~\cite{endosurf}, extending EndoNeRF~\cite{EndoNeRF} and showing improved performance in surface estimation. These approaches leverage \gls{sdf} formulations, allowing geometry to be disentangled and additional geometric cues to regularize the reconstructed surface.

~\citet{lightneus} develop LightNeuS for endoscopic surface reconstruction using monocular videos. They identify two properties of endoscopic videos. First, the surface can be considered tubular and therefore easily modeled by \glspl{sdf}; second, the light characteristics of endoluminal cavities can be approximated by an inverse-squared law. This allows the rendering equation to be rewritten in terms of depth. Reconstruction improves, but this method lacks the ability to handle soft deformation.

\Gls{nerf} can also be used for imaging modalities that are intrinsically 3D such as Magnetic Resonance Imaging (MRI), and Computed Tomography (CT). 
CuNeRF~\cite{cunerf} generates high-resolution MRI and CT images from low-resolution inputs using NeRF formulations. Fang et al.~\cite{fang2022curvature} employ NeRF-style implicit functions to enhance resolution and accuracy in 3D tooth reconstruction from Cone Beam CT scans. 
In another example, \cite{wiesner2022implicit} shows the generalizability of \gls{sdf} representations for generating models of living cell shapes (microscopy imaging). They show how additional input parameters such as time and a latent code can condition the representation to predict the evolution of the cell shapes. 

Assessing the effectiveness of emerging technologies like neural rendering in medical applications is essential for medical research. In the context of chronic wound documentation, Syn3DWound~\cite{lebrat2023syn3dwound} assesses a \gls{nerf}-based model for 3D wound reconstruction from synthetic images, comparing its performance to well-established photogrammetry solutions. 
\subsection{Digital Human Reconstruction}
\label{sec4:digital_humans}

3D human reconstruction involves challenges ranging from motion to non-rigid deformation, often found in tasks of inverse kinematics and skeletal animation. Unlike for the reconstruction of arbitrary scenes, the human body can be represented with a mesh template such as the Skinned Multi-Person Linear model (SMPL)~\cite{smpl2023}, which can accurately model pose-dependent shape variations.
However, a central disadvantage of this model is its limited capacity to represent fine details such as clothing and hairstyle~\cite{jia2024human}. 

In this direction, several works integrated NeRF with the scope of rendering photorealistic images. Unlike standard NeRF settings, 3D human reconstruction typically involves monocular video of dynamic subjects. 
One approach to handling this is to encode temporal changes alongside the standard three dimensions in a unified 4D space.
Neural Body~\cite{Peng_2021_CVPR} encodes structured latent codes anchored to an SMPL model into a single neural radiance field, providing implicit geometric guidance.
Alternatively, other methods optimize a 3D canonical space from pose-estimation priors~\cite{Weng_2022_CVPR}, extending Dynamic Scenes (see Section 3.6.1) to target anatomical structures, such as hands~\cite{Guo_2023_CVPR} and heads~\cite{Hong_2022_CVPR}, and non-rigid deformations regularizing deformations across time~\cite{Park2021Nerfies,park2021hypernerf}.

To enforce multi-view geometric correspondence of finer details, ~\cite{gao2023neural,zhou2024animatable} utilize human masks to optimize an implicit SDF field, preserving high-frequency details at image boundaries. ~\cite{xiu2022icon} further addresses visibility challenges by leveraging the visibility-aware occupancy field of the learned geometry, enabling more precise surface reconstruction even in scenarios with significant occlusion.
Building on this idea, ~\cite{wang2023adaptive} employs a hybrid volumetric-surface method by fitting explicit shells near learned surfaces, drastically accelerating rendering while maintaining geometric accuracy. This approach proves particularly effective for modeling complex parts of the human head, where the head is represented by an SDF and the hair structure is captured by a volumetric field.

While previous methods primarily focus on reconstructing geometry, generation-based techniques aim to synthesize novel appearances. For instance, ~\cite{OrEl_2022_CVPR} combines an SDF with a 2D style generator to synthesize novel photorealistic appearances. In contrast, ~\cite{Jiang_2023_CVPR} leverages explicit reconstruction priors derived from a 2D human generator to learn accurate and detailed 3D human geometry.
For a more detailed overview of these techniques, in particular for methods focusing on human generation, we refer the reader to~\cite{jia2024human,wang2024survey}.

\section{Computer Vision Tasks beyond Reconstruction}
\label{sec:computer_vision}

\Gls{nerf} has emerged as a foundational 3D scene representation and has powered a recent surge in applications beyond 3D reconstruction.
This section highlights the significance of \glspl{nerf} in three common applications: robotics-related applications (Section~\ref{sec5:robotics}) and traditional computer vision tasks (Section~\ref{sec5:recognition}).
We further provide a concise review of 3D generation and editing (Section~\ref{sec5:generative_ai}).

\subsection{Applications on Robotics}
\label{sec5:robotics}

This subsection discusses \gls{nerf}'s capabilities for \gls{slam} — a fundamental and long-established component of robotic perception. 
It simultaneously maps an unknown environment while tracking the robot's location, which inherits challenges associated with camera pose estimation, 3D reconstruction, and robustness to dynamic or transient scene elements.
Traditional dense \gls{slam} methods generally utilize features extracted from RGB images, depth maps, or LiDAR scans and often rely on voxel-based scene representations (e.g., occupancy grids or TSDFs). While effective for dense mapping, these representations typically exhibit a trade-off between reconstruction accuracy and memory consumption.
Moreover, many conventional approaches depend on carefully engineered modules and hand-crafted optimization pipelines, which limit their flexibility and ability to benefit from end-to-end learning frameworks.

These limitations have motivated the integration of \glspl{inr}, in particular NeRF, into SLAM pipelines, where
iMAP~\cite{sucar2021imap} represents one of the earliest efforts.
\Gls{nerf} enhances \gls{slam} by providing a continuous, differentiable scene representation from 2D images.
This implicit representation enables compact scene encoding with bounded memory requirements, while its differentiable volume rendering formulation supports incremental test-time optimization, improving both reconstruction quality and tracking accuracy.
NICE-SLAM~\cite{zhu2022nice} addresses scalability and detail preservation in large-scale \gls{slam} by using hierarchical, grid-based neural encoding with multi-level local information and pre-trained geometric priors, allowing for efficient local updates. 
NeRF-SLAM~\cite{rosinol2023nerf} improves 3D mapping and scene reconstruction from monocular images, enhancing geometric and photometric accuracy through accurate pose and depth estimations. 
Dense RGB SLAM~\cite{li2023dense} employs a hierarchical feature volume to improve accuracy in large scenes and incorporates a multi-scale patch-based warping loss for a robust dense visual \gls{slam} system by constraining camera motion and the scene map. 
Subsequent research explores point-based (Point-SLAM~\cite{sandstrom2023pointslam}), grid-based (ESLAM~\cite{johari2023eslam}), and hybrid (Co-SLAM~\cite{wang2023coslam}) representations to alleviate the trade-off between representation efficiency and reconstruction and tracking performance.
For example, Point-SLAM utilizes a point-based representation, which enables adaptive point allocation according to local information density.
Loopy-SLAM~\cite{liso2024loopy} further incorporates loop closure into neural \gls{slam} to mitigate error accumulation during tracking, ensuring global consistency through pose graph optimization and online loop closure detection.
In addition, iSDF~\cite{ortiz2022isdf} enables incremental updates to \gls{sdf}, allowing \gls{slam} systems to adapt more rapidly to environmental change for high-resolution mapping.
Despite these advances in \gls{nerf}-based \gls{slam}, several challenges remain. These include balancing memory consumption, computational cost, and reconstruction and tracking performance, as well as handling the forgetting problem and developing effective keyframe selection strategies.

%
Beyond the \gls{slam} task, \gls{nerf} offers a high-fidelity 3D scene representation that provides richer geometry and supports 3D scene understanding beyond individual 2D images, thereby improving performance in trajectory planning and complex control tasks (e.g., object grasping).


Path planning is essential for autonomous robotic navigation, ensuring collision-free trajectories. \Gls{nerf} improves path planning by offering high-fidelity 3D representations from 2D images, enabling detailed geometric understanding and smoother trajectory planning in dynamic environments.
NeRF-Nav~\cite{adamkiewicz2022vision} utilizes \glspl{nerf} to create detailed geometric maps, optimizing trajectories by minimizing collision risks through density fields. It also incorporates a vision-based state estimator for real-time dynamic replanning and robust navigation.
RNR-Map~\cite{kwon2023renderable} integrates visual and semantic cues for efficient mapping and localization, demonstrating highly accurate and robust navigation even in changing environments. 
CATNIPS~\cite{chen2024catnips} transforms \glspl{nerf} into Poisson Point Processes to quantify collision probabilities. It integrates a voxel-based probabilistic unsafe robot region to ensure safe path planning.
NeRF2Real~\cite{byravan2023nerf2real} transfers vision-based bipedal locomotion policies from simulation to reality, leveraging NeRF to generate photorealistic 3D scenes combined with physics simulations to train reinforcement learning policies. 
MultiOb~\cite{marza2023multi} introduces two neural implicit representations for agent navigation: one for predicting object positions and the other for encoding occupancy and exploration, significantly enhancing navigation capabilities in complex environments. 
Integrating NeRFs into path planning can support autonomous navigation by improving 3D mapping, collision avoidance, trajectory optimization, and semantic understanding. 
Future efforts should optimize computational efficiency for real-time use and integrate NeRFs with sensors such as LiDAR and radar for greater effectiveness. 

Robotic control must support high-level tasks such as path planning and perception in complex environments containing challenging objects.
NeRF’s high-fidelity 3D representations enhance robotic systems by providing detailed data for improved control.
For example, NeRF-CBF~\cite{tong2023enforcing} uses NeRFs for single-step visual foresight to support control barrier functions and enhance safety and reliability through future observation predictions. Real-time simulations show its superior performance in preventing unsafe actions.
Controlling3D~\cite{li20223d} connects visual perception with motor control, generating detailed 3D environment representations that improve the accuracy and robustness of visuomotor control in dynamic and unstructured environments.
NeRF-supervision~\cite{yen2022nerf} introduces a self-supervised pipeline to generate accurate dense correspondences, enhancing visual descriptors for challenging objects, such as thin and reflective surfaces. 
Evo-NeRF~\cite{kerr2022evo} improves robotic grasping efficiency for transparent objects by combining rapid NeRF training with sequential adaptation, allowing quick updates to the scene representation as objects are removed.
GraspNeRF~\cite{dai2023graspnerf} employs a multi-view RGB-based 6-DoF grasp detection network with generalizable NeRF for real-time material-agnostic object grasping, including transparent and specular objects in cluttered environments. 
To improve object manipulation and navigation, SPARTN~\cite{zhou2023nerf} uses an offline data augmentation technique with NeRF to inject corrective noise into visual demonstrations, enabling real-time, RGB-only 6-DoF grasping policies that improve grasp success rates.
NGDF~\cite{weng2023neural} models grasping as the level set of a continuous implicit function, enabling joint optimization of grasping and motion planning through gradient-based trajectory optimization.
NeRF’s powerful 3D scene representation may offer useful capabilities for robots operating around challenging objects and in complex environments. However, achieving real-time performance remains a major challenge for effective interaction with the real world.

\subsection{Recognition Tasks}
\label{sec5:recognition}

Visual recognition is a fundamental problem in computer vision, involving the identification of semantically meaningful entities within visual data. This encompasses a wide spectrum of tasks, including but not limited to semantic segmentation, object detection, and object tracking. 
While current methods are effective on 2D data such as images and video frames~\cite{cheng2021mask2former}, they heavily rely on pre-trained architectures fine-tuned on large annotated 2D datasets. Extending these methods to 3D data (e.g. point cloud and triangular mesh) is challenging due to the scarcity of 3D annotations, as annotating in 3D is labor-intensive. Methods that use only 2D annotations struggle to integrate 2D knowledge into accurate 3D representations, often resulting in errors.

In this context, NeRF can integrate multiple 2D observations into a unified, geometrically and semantically consistent 3D model, reducing the need for extensive 3D annotations.
SemanticNeRF~\cite{zhi2021place} performs 3D semantic segmentation by introducing a NeRF model that jointly learns scene appearance, semantics, and geometry from annotated 2D images. 
This work extends the original NeRF model by including a view-invariant function that maps world coordinates to semantic labels, enabling the rendering of precise 2D segmentation through the conventional NeRF's rendering equations. It can perform 2D-to-3D label propagation, even with sparse, noisy, and low-resolution 2D annotations. 
Similarly, works like~\cite{fu2022panoptic,Siddiqui_2023_CVPR} achieve 3D panoptic segmentation by using coarse 3D annotations, such as 3D bounding boxes, and designing a label propagation scheme to handle ambiguities and noise.

Targeting the challenging task of multimodal open-set recognition, LERF~\cite{lerf2023} integrates CLIP embeddings~\cite{radford2021learning} into NeRF to enable natural-language-based search and localization of objects, visual attributes, and geometric details within 3D scenes. 
SA3D~\cite{NEURIPS2023_525d2440} \textit{lifts} 2D vision foundation models like Segment Anything (SAM)~\cite{kirillov2023segany} to 3D, providing an interactive 3D segmentation through simple user prompts.
NeRFs can also leverage scene physics and rendering effects as self-supervision signals to learn visual recognition models without relying on 2D or 3D annotations. For instance, D$^2$-NeRF~\cite{wu2022d2nerf} segments moving objects by decoupling the static and dynamic scene elements.

While the previous methods are scene-specific and depend on reconstruction quality, other approaches aim for a general model to perform visual recognition across multiple NeRF-encoded scenes. 
Trained NeRFs can act as a versatile visual data format, capable of generating 2D images from various viewpoints or creating sequences that simulate video.
Building on this, some methods perform visual recognition within the NeRF's implicit scene representation. For instance, NeRF-RPN~\cite{nerfrpn} detects 3D objects in a NeRF scene by training a standard object detector on sampled NeRF primitives (\textit{e.g.},~density and color).  NeSF~\cite{vora2022nesf} achieves semantic segmentation across scenes by training a neural network to map sampled NeRF representations to generic features, which can then be segmented by deep-learning segmentation models.

A crucial aspect of applying NeRFs to visual recognition tasks lies in the interplay between the geometric and semantic features of a scene. Often, one aims to exploit geometric features to semantically distinguish objects within a scene. However, research has also demonstrated that integrating semantic reasoning into NeRF models can contribute to a more comprehensive understanding of scene geometry and appearance, thereby improving the model's efficacy for novel view synthesis. 
For instance, DietNeRF~\cite{Jain_2021_ICCV} and SinNeRF~\cite{xu2022sinnerf} utilize semantic reasoning to address data deficiencies such as limited view diversity and inaccurate camera poses, boosting generalization across unseen viewpoints and environments.
Yet, these methods usually fail to predict fine-grained details on unobserved areas.

\subsection{3D Generation and Editing}
\label{sec5:generative_ai}

Recently, generative AI techniques such as Stable Diffusion~\cite{rombach2022stablediffusion} and DALL-E~\cite{ramesh2022dalle} have achieved impressive results in text-to-image generation and image editing.
Extending those generative capabilities from 2D to 3D content opens new opportunities for entertainment like video games and creative industries like immersive experiences, where rich 3D assets are essential but costly to craft manually. 
Nevertheless, lifting the generation to 3D is not as simple as it seems.
It requires 1) proper scene representation, and 2) semantic alignment between the input text descriptions and the generated 3D content.
\gls{nerf}~\cite{Mildenhall2020NeRF} has been a popular choice for scene representations as its adopted volume rendering allows for consistent generation of 3D contents from 2D images.

For semantics alignment, early works like DreamFields~\cite{jain2022dreamfields} aim to minimize the CLIP distance~\cite{radford2021clip} between rendered views and input text prompts.
However, the use of pre-trained 2D image-text models compromises realism and accuracy in the generated 3D content. 
To address this, DreamFusion~\cite{poole2023dreamfusion} extends the generative capabilities of 2D diffusion models~\cite{saharia2022photorealistic} to 3D.
Specifically, it replaces the CLIP loss with a loss derived from the distillation of a 2D diffusion model, introducing the Score Distillation Sampling (SDS) framework. This enhancement improves the quality and realism of the generated 3D content.
%
Expanding on this foundational idea, various architectural improvements have been proposed to enhance quality and computational efficiency.
Some approaches initialize NeRF with a 3D shape prior~\cite{xu2023dream3d}, or a 3D point cloud~\cite{yu2023points_to_3d}.
Magic3D~\cite{lin2023magic3d} introduces a coarse-to-fine reconstruction strategy using Instant NGP~\cite{instant-ngp} alongside mesh extraction and differentiable mesh rendering. 

Images can also inform 3D generation tasks by explicitly providing geometry and appearance information of 3D contents.
The image-to-3D task focuses on inferring the 3D structures and appearance of objects from single or sparse images.
Some research~\cite{tang2023makeit3d, melas2023realfusion} derives or optimizes text prompts from the input image and utilizes the DreamFusion-like scheme to learn the 3D shape.
However, these methods are time-consuming and often suffer from the 3D inconsistency problem as the adopted 2D diffusion models fail to provide view-consistent supervision.
Differently, Zero-1-to-3~\cite{liu2023zero123} designs a view-conditioned diffusion model from a single image for 3D-consistent novel view synthesis and then lifts these synthesized views for 3D shape learning.
Based on this scheme, One-2-3-45~\cite{liu2024one2345} further enhances the 3D generation performance and efficiency with a few-shot SDF-based model (SparseNeuS~\cite{long2022sparseneus}), leading to a 3D mesh generation from any single image in 45 seconds.

The generative properties of these models can be further tailored for controllable 3D content generation.
Related works focus on incorporating multiple appearance or geometrical constraints in 3D generation.
For instance, DreamBooth3D~\cite{raj2023dreambooth3d} constrains text-to-3D generation using a few input images. Similarly,  Control3D~\cite{chen2023control3d} directs geometry through an input sketch, while Set-the-Scene~\cite{cohen2023set-the-scene} allows 3D scene generation with object proxies determining the placement of different elements. 
LaTeRF~\cite{mirzaei2022laterf} extracts 3D assets from a pre-trained \gls{nerf} model, guided by a few pixel annotations and a natural language description of the target object.
The ability to finetune 3D-generated content is another valuable feature of these methods. 
Magic3D~\cite{lin2023magic3d} proposes prompt-based editing through fine-tuning, whereas DreamBooth3D~\cite{raj2023dreambooth3d} allows 3D object generation with simple prompt adjustments such as re-contextualization, accessorization, or stylization.

While major works revolve around single-object or dream-style generation, research interest grows towards expanding this generative capability to larger and more realistic environments.
For example, Text2NeRF~\cite{zhang2024text2nerf} explores the use of progressive in-painting and depth supervision to enhance the realism of 3D scene generation.
The generative potential of these approaches has also sparked wide interest in digital human creation. Its applications range from 3D human generation~\cite{xiong2023get3dhuman} and 3D face model generation~\cite{chan2022eg3d}, to talking-head synthesis~\cite{guo2021adnerf} and face morphing~\cite{wang2022morf}.

In addition to generation, \gls{nerf}-based 3D \acrshort{aigc} techniques show promise in scene editing.
NeRF-Editing~\cite{yuan2022nerf} enables users to manipulate scene geometry by editing the mesh extracted from a learned \gls{nerf}.
Given a text prompt or an exemplar image, one can modify the local properties of an object, such as  appearance~\cite{wang2022clip, kuang2023palettenerf} and geometry~\cite{wang2022clip, jambon2023nerfshop}, or manipulate objects within a scene~\cite{zhuang2023dreameditor}.
Other works focus on altering a scene's global properties while keeping its original structure, performing relighting (see Section 3.4) or stylization~\cite{liu2023stylerf}. 
Recently, Instruct-NeRF2NeRF~\cite{haque2023instructnerf2nerf} has shown compelling editing ability in both local (\textit{e.g.}~appearance) and global (\textit{e.g.}~style) modification using text prompts.

An important avenue of future research on 3D generation techniques includes extending related advancements to dynamic content, as explored by MAV3D~\cite{singer2023textto4d}, and further reducing the generation time through amortized optimization as demonstrated by ATT3D~\cite{lorraine2023att3d}.
We refer our readers to recent surveys~\cite{li2024aigc_survey, zhan2023multimodal} for a detailed discussion on recent advancements of 3D generation techniques and their respective applications.

\section{Tools And Datasets}
\label{sec:practice}

This section compiles a list of tools to facilitate the development and analysis of \glspl{nerf}. We also provide commonly used datasets and evaluation protocols for benchmarking.

\subsection{Tools}
\label{sec6:tools}

This subsection introduces tools related to data generation, data preprocessing, NeRF training and optimization, and visualization/rendering.

\subsubsection{Dataset generation}
Synthetic datasets are usually created with fully controllable variables, making them a crucial first step in developing effective NeRF algorithms. Tools commonly used to create synthetic datasets for NeRF include, but are not limited to, Blender~\cite{blender} and Unity~\cite{unity}. 
Blender, an open-source 3D computer graphics software, is widely used for synthetic dataset generation, such as D-NeRF~\cite{d-nerf}, by simulating various environments and lighting conditions. BlenderNeRF~\cite{Raafat_BlenderNeRF_2024} integrates with Blender to streamline the process of generating high-quality and diverse datasets for novel-view synthesis tasks. Unity, a game engine, can also create synthetic data through custom scripts and scenarios. Unity NeRF\footnote{\url{https://github.com/kwea123/nerf_Unity}. Accessed: 2024-11-05.} is developed specifically for generating NeRF datasets within the Unity environment.

\subsubsection{Data preprocessing}
Most NeRF approaches require camera poses as priors, calculating camera poses is an important step in preprocessing multi-view images. Several tools are available for this purpose, including COLMAP~\cite{schonberger2016structure}, pixSfM~\cite{lindenberger2021pixel}, and OpenDroneMap~\cite{gbagir2023opendronemap}.
COLMAP is widely used for 3D reconstruction and preprocessing image data, providing accurate camera poses and point clouds. 
PixSfM improves the accuracy of COLMAP by refining 3D points and camera poses with direct alignment of 2D image features extracted from deep-learning methods. OpenDroneMap is useful for preprocessing large-scale datasets, allowing rapid 3D reconstruction with the possibility of including image GPS coordinates, camera yaw, pitch, and roll to accelerate SfM.

\subsubsection{Training and optimization}
To standardize the training and optimization of NeRF models and facilitate user-friendly interaction with ~\gls{nerf} models,  NeRFStudio~\cite{nerfstudio} provides an easy-to-use API for modular NeRF development. It simplifies model architecture design, training, and hyperparameter tuning, facilitating efficient development and optimization of NeRF models. 
Building on NeRFStudio, SDFStudio~\cite{Yu2022SDFStudio} focuses on 3D reconstruction and provides a unified and modular framework for neural implicit surface reconstruction. 
Focusing on efficient sampling in NeRF, NerfAcc~\cite{Li2023NerfAcc:NeRFs} offers a PyTorch \gls{nerf} acceleration toolbox designed for both training and inference. It is a universal and plug-and-play tool compatible with most NeRF models.

\subsection{Datasets and Evaluation Protocols}
\label{sec6:datasets}
{

This section provides an overview of widely used datasets in \gls{nerf}-based research, as summarized in Table~\ref{tab:datasets}. It first reviews benchmarking datasets and evaluation metrics for two fundamental tasks --- \emph{novel view synthesis} and \emph{3D surface reconstruction}, along with representative datasets and their corresponding leading methods provided in Table~\ref{tab:dataset_sota}. 
It then discusses datasets that facilitate the broad adoption of \glspl{nerf} in real-world applications.

\definecolor{outdoors}{HTML}{c77dff}
\definecolor{Object}{HTML}{D85531}
\definecolor{Indoors}{HTML}{197BBD}
\definecolor{Static}{HTML}{3BB273}
\definecolor{Dynamic}{HTML}{424B54}
\newcommand{\out}{\textcolor{outdoors}{O}}
\newcommand{\obj}{\textcolor{Object}{J}}
\newcommand{\indo}{\textcolor{Indoors}{I}}
\newcommand{\static}{\textcolor{Static}{S}}
\newcommand{\dynamic}{\textcolor{Dynamic}{D}}
\renewcommand{\faPlus}{\wendCheck}
\renewcommand{\faMinus}{\nope}

\begin{table}[!t]\centering
\caption{Commonly used datasets for \glspl{nerf}-related methods. In the column of \textbf{Content}, \obj~means \textcolor{Object}{objects}, \indo~ means \textcolor{Indoors}{indoor scenes}, and \out~means \textcolor{outdoors}{outdoor scenes}. In the column of \textbf{Type}, \static~means \textcolor{Static}{static scenes} and \dynamic~means \textcolor{Dynamic}{dynamic scenes}.}\label{tab:datasets}
\scriptsize
\adjustbox{max width=\textwidth}{
\renewcommand{\arraystretch}{2}
\begin{tabular}{r|l|c|c|l|c|c|c|c|c|ccc|c}\toprule
\multirow{2}{*}{Scene}  
&\multirow{2}{*}{Name} 
&\multirow{2}{*}{Content}
&\multirow{2}{*}{Type}
&\multirow{2}{*}{Scene Num./Size} 
&\multicolumn{5}{c|}{Max. Resolution (Width, Height)} 
&\multirow{2}{*}{Depth}
&\multirow{2}{*}{Semantics}
&\multirow{2}{*}{3D}
&\multirow{2}{*}{Website} \\ \cline{6-10}
& & & & &\ [0, 1k)\  &\ [1k, 2k) &\ [2k, 3k) &\ [3k, 4k) & [4k, $+\infty$) & & & & \\\cline{1-14}
%
\multirow{9}{*}{Synthetic}  
&NeRF Synthetic~\cite{Mildenhall2020NeRF}
&\obj           &\static
&\makecell[l]{8 scenes\\  100 training \\ 200 testing}. 
&\faPlus        &\faMinus       &\faMinus       &\faMinus       &\faMinus 
&\wendCheck     &\nope          &\nope 
&\href{https://www.matthewtancik.com/nerf}{\faLink}
\\\cline{2-14}
&ShapeNetCore\tablefootnote{ShapeNetCore dataset does not provide rendered images. A common way is to follow 3D-R2N2~\cite{choy20163dr2n2} to render images and split the dataset.}~\cite{chang2015shapenet}
&\obj           &\static
&\makecell[l]{51300 3D models \\ 55 object categories} 
& & & & &
&\nope          &\nope          &\wendCheck 
&\href{https://shapenet.org/}{\faLink}
\\\cline{2-14}
&Shiny Blender~\cite{ref-nerf}  
&\obj           &\static
&\makecell[l]{6 scenes \\  100 training \\ 200 testing} 
&\faPlus        &\faMinus       &\faMinus       &\faMinus       &\faMinus 
&\wendCheck     &\nope          &\nope 
&\href{https://dorverbin.github.io/refnerf/}{\faLink}
\\\cline{2-14}
&D-NeRF~\cite{d-nerf}         
&\obj           &\dynamic
&\makecell[l]{8 scenes \\  100 training \\ 200 testing} 
&\faPlus        &\faMinus       &\faMinus       &\faMinus       &\faMinus 
&\wendCheck     &\nope          &\nope 
&\href{https://www.albertpumarola.com/research/D-NeRF/index.html}{\faLink}
\\\cline{2-14}
&Objaverse\tablefootnote{Objaverse provides \href{https://github.com/allenai/objaverse-xl}{scripts} for rendering images with Blender.}~\cite{deitke2023objaverse}
&\obj           &\static~\&~\dynamic
&\makecell[l]{818K objects \\21K classes}
&               &               &               &               &
&\nope          &\nope          &\wendCheck
&\href{https://objaverse.allenai.org/}{\faLink}
\\\cline{2-14}
&BlendedMVS~\cite{yao2020blendedmvs}     
&\obj \ \& \out &\static
& \makecell[l]{113 scenes \\  20 to 1K images} 
&\faPlus        &\faMinus       &\faPlus        &\faMinus       &\faMinus 
&\wendCheck     &\nope          &\wendCheck 
&\href{https://github.com/YoYo000/BlendedMVS}{\faLink}
\\\cline{2-14}
&MatrixCity~\cite{li2023matrixcity}     
&\out           &\dynamic
&\makecell[l]{2 scenes \\ 67k aerial images \\ 452k street-level images} 
&\faMinus       &\faPlus        &\faMinus       &\faMinus       &\faMinus 
&\wendCheck     &\nope          &\nope 
&\href{https://city-super.github.io/matrixcity/}{\faLink}
\\\cline{2-14}
&Deblur-NeRF~\cite{ma2022deblurnerf}
&\out\ \& \indo &\static
&\makecell[l]{10 scenes \\ 34 images} 
&\faPlus        &\faMinus       &\faMinus       &\faMinus       &\faMinus 
&\nope          &\nope          &\nope 
&\href{https://limacv.github.io/deblurnerf/}{\faLink}
\\\cline{2-14}
&Replica\tablefootnote{Replica originally contains the 3D model of scenes. One can render datasets using their provided engines.}~\cite{replica19arxiv}  
&\indo          &\static
&\makecell[l]{18 scenes} 
& & & & &
&\wendCheck     &\wendCheck     &\wendCheck 
&\href{https://github.com/facebookresearch/Replica-Dataset}{\faLink}
\\\cline{1-14}
\multirow{23}{*}{Real}
&Deblur-NeRF~\cite{ma2022deblurnerf}    
&\obj           &\static
&\makecell[l]{21 scenes \\ 27 to 52 images} 
&\faPlus        &\faMinus       &\faPlus        &\faMinus       &\faMinus
&\nope          &\nope          &\nope 
&\href{https://limacv.github.io/deblurnerf/}{\faLink}
\\ \cline{2-14}
&Shiny dataset~\cite{wizadwongsa2021shiny}  
&\obj           &\static
&\makecell[l]{8 scenes \\  32 to 307 images} 
&\faMinus       &\faPlus        &\faMinus       &\faMinus       &\faMinus 
&\nope          &\nope          &\wendCheck 
&\href{https://nex-mpi.github.io/}{\faLink}
\\\cline{2-14}
&DyNeRF~\cite{li2022neural}         
&\obj           &\dynamic
&\makecell[l]{6 scenes \\ 10 seconds video (30 fps)} 
&\faMinus       &\faMinus       &\faMinus       &\faPlus        &\faMinus 
&\nope          &\nope          &\nope 
&\href{https://neural-3d-video.github.io/}{\faLink}
\\\cline{2-14}
&HyperNeRF~\cite{park2021hypernerf}      
&\obj           &\dynamic
&\makecell[l]{17 scenes \\ 82 to 415 images} 
&\faPlus        &\faPlus        &\faMinus       &\faMinus       &\faMinus 
&\nope          &\nope          &\nope 
&\href{https://hypernerf.github.io/}{\faLink}
\\\cline{2-14}
&Nerfies~\cite{Park2021Nerfies}        
&\obj           &\dynamic
&\makecell[l]{9 scenes \\ 40 to 356 images} 
&\faPlus        &\faPlus        &\faMinus       &\faMinus       &\faMinus 
&\nope          &\nope          &\nope 
&\href{https://nerfies.github.io/}{\faLink}
\\\cline{2-14}
&DTU~\cite{aanaes2016dtu}            
&\obj           &\static
&\makecell[l]{80 scenes \\ 49 to 64 images} 
&\faMinus       &\faPlus        &\faMinus       &\faMinus       &\faMinus 
&\nope          &\nope          &\wendCheck 
&\href{https://roboimagedata.compute.dtu.dk/?page_id=36}{\faLink}
\\\cline{2-14}
&CO3D~\cite{reizenstein2021co3d}          
&\obj           &\static
&\makecell[l]{19K objects \\50 categories \\1.5M images} 
&\faPlus        &\faMinus        &\faMinus       &\faMinus       &\faMinus 
&\wendCheck     &\nope          &\wendCheck 
&\href{https://github.com/facebookresearch/co3d}{\faLink}
\\\cline{2-14}
&GSO\tablefootnote{GSO dataset does not provide images. One can follow Zero-1-to-3~\cite{liu2023zero123} to generate rendering.}~\cite{downs2022gso}           
&\obj           &\static
&\makecell[l]{1030 objects \\17 categories} 
&               &               &              &               & 
&\nope          &\nope          &\wendCheck 
&\href{https://app.gazebosim.org/GoogleResearch}{\faLink}
\\\cline{2-14}
&OmniObject3D~\cite{wu2023omniobject3d}           
&\obj           &\static
&\makecell[l]{6000 objects \\ 190 categories} 
&\faPlus        &\faPlus        &\faMinus       &\faMinus       &\faMinus 
&\wendCheck     &\nope          &\wendCheck 
&\href{https://omniobject3d.github.io/}{\faLink}
\\\cline{2-14}
&ScanNet~\cite{dai2017scannet}        
&\indo          &\static
&\makecell[l]{1503 scenes}
&\faMinus       &\faPlus        &\faMinus       &\faMinus       &\faMinus 
&\wendCheck     &\wendCheck     &\wendCheck 
&\href{https://www.scan-net.org/}{\faLink}
\\\cline{2-14}
&ScanNet++~\cite{yeshwanth2023scannet++}      
&\indo          &\static
&\makecell[l]{460 scenes \\ 280000 DSLR images \\ 3.7M RGB-D frames} 
&\faMinus       &\faPlus        &\faMinus       &\faMinus       &\faPlus  
&\wendCheck     &\wendCheck     &\wendCheck 
&\href{https://kaldir.vc.in.tum.de/scannetpp/}{\faLink}
\\\cline{2-14}
&RealEstate10K~\cite{realestate10k}  
&\indo          &\static
&\makecell[l]{1500 scenes \\ around 750K frames} 
&\faMinus       &\faPlus        &\faMinus       &\faMinus       &\faMinus 
&\nope          &\nope          &\nope 
&\href{https://google.github.io/realestate10k/}{\faLink}
\\\cline{2-14}
&ARKitScenes~\cite{arkitscenes}    
&\indo          &\static
&\makecell[l]{1661 scenes \\ 5047 scans} 
&\faPlus        &\faPlus        &\faMinus       &\faMinus       &\faMinus 
&\wendCheck     &\wendCheck     &\wendCheck 
&\href{https://github.com/apple/ARKitScenes}{\faLink}
\\\cline{2-14}
&EyefulTower~\cite{xu2023VRNeRF}    
&\indo          &\static
&\makecell[l]{13 scenes \\ 765 to 8030 images} 
&\faMinus       &\faPlus        &\faPlus        &\faPlus        &\faPlus  
&\nope          &\nope          &\wendCheck 
&\href{https://github.com/facebookresearch/EyefulTower}{\faLink}
\\\cline{2-14}
&Dl3dv-10k~\cite{ling2024dl3dv}       
&\indo\ \& \out  &\static
&\makecell[l]{10510 videos \\ 51.2M frames} 
&\faPlus        &\faMinus       &\faPlus        &\faPlus        &\faMinus  
&\nope          &\nope          &\nope 
&\href{https://dl3dv-10k.github.io/DL3DV-10K/}{\faLink}
\\\cline{2-14}
&ETH3D~\cite{schops2017eth3d_multi}       
&\indo\ \& \out  &\static
&\makecell[l]{25 scenes \\7 to 110 images} 
&\faMinus       &\faMinus       &\faMinus       &\faMinus       &\faPlus  
&\wendCheck     &\nope          &\wendCheck 
&\href{https://www.eth3d.net/datasets#high-res-multi-view}{\faLink}
\\\cline{2-14}
&LLFF-Real~\cite{mildenhall2019llff}      
&\indo\ \& \out &\static
&\makecell[l]{24 scenes \\ 20 to 30 images} 
&\faMinus       &\faMinus       &\faMinus       &\faMinus       &\faPlus  
&\nope          &\nope          &\nope 
&\href{https://github.com/Fyusion/LLFF}{\faLink}
\\\cline{2-14}
&TanksAndTemples\tablefootnote{Most NeRF-related methods use these two variants of pre-processed datasets---Plenoxels~\cite{Fridovich2022Plenoxels} (with backgrounds) and NSVF~\cite{Liu2020NSVF} (without backgrounds).}~\cite{Knapitsch2017tnt}
&\indo\ \& \out &\static
&\makecell[l]{14 scenes} 
&\faMinus       &\faPlus        &\faPlus        &\faPlus        &\faPlus  
&\nope          &\nope          &\wendCheck 
&\href{https://www.tanksandtemples.org/}{\faLink}
\\\cline{2-14}
&MipNeRF-360~\cite{Barron2022MipNeRF360}    
&\indo\ \& \out &\static
&\makecell[l]{9 scenes \\ 100 to 330 images} 
&\faPlus        &\faPlus        &\faPlus        &\faMinus       &\faPlus  
&\nope          &\nope          &\nope 
&\href{https://jonbarron.info/mipnerf360/}{\faLink}
\\\cline{2-14}
&Phototourism~\cite{jin2021phototourism}   
&\out\          &\static
&\makecell[l]{25 scenes \\ 75 to 3765 images} 
&\faMinus       &\faPlus        &\faMinus       &\faMinus       &\faMinus 
&\wendCheck     &\nope          &\wendCheck 
&\href{https://phototour.cs.washington.edu/datasets/}{\faLink}
\\\cline{2-14}
&Mill 19~\cite{Turki2022Mega-NeRF}        
&\out           &\dynamic
&\makecell[l]{2 scenes \\ 1678 and 1940 images} 
&\faMinus       &\faMinus       &\faMinus       &\faMinus       &\faPlus  
&\nope          &\nope          &\nope 
&\href{https://github.com/cmusatyalab/mega-nerf}{\faLink}
\\\cline{2-14}
&Kitti-360~\cite{liao2022kitti360}       
&\out           &\dynamic
&\makecell[l]{11 scenes \\ 320K images \\ 80K laser scans} 
&\faMinus       &\faPlus        &\faMinus       &\faMinus       &\faMinus  
&\nope          &\wendCheck     &\wendCheck 
&\href{https://www.cvlibs.net/datasets/kitti-360/}{\faLink}
\\\cline{2-14}
&Waymo Block-NeRF~\cite{Tancik2022BlockNeRF}       
&\out           &\dynamic
&\makecell[l]{12000 images} 
&\faMinus       &\faPlus        &\faMinus       &\faMinus       &\faMinus  
&\nope          &\nope          &\nope 
&\href{https://waymo.com/research/block-nerf}{\faLink}
\\ \cline{1-14}
%
\multirow{2}{*}{\makecell[r]{Real \& \\Synthetic}}
&UrbanScene3D~\cite{lin2022urbanscene3d}   
&\out           &\static
&\makecell[l]{10 synthetic \\ 6 real \\ 128K images} 
&\faMinus       &\faMinus       &\faMinus       &\faMinus       &\faPlus  
&\wendCheck     &\wendCheck     &\wendCheck 
&\href{https://vcc.tech/UrbanScene3D}{\faLink}
\\\cline{2-14}
&Objaverse-XL\tablefootnote{Objaverse-XL dataset provides \href{https://github.com/allenai/objaverse-xl}{scripts} for rendering images with Blender.}~\cite{deitke2023objaversexl}
&\obj           &\static~\&~\dynamic
&\makecell[l]{10.2M objects}
&               &               &               &               &
&\nope          &\nope          &\wendCheck
&\href{https://objaverse.allenai.org/}{\faLink}
\\\bottomrule
\end{tabular}
}
\end{table}

\paragraph{Novel View Synthesis.}
A typical dataset is the synthetic dataset from the original \gls{nerf} paper~\cite{Mildenhall2020NeRF}, NeRF Synthetic. It consists of eight $360^\circ$ object-centric scenes covering different levels of geometric complexities (e.g., reflection, occlusion, and thin details) and offers accurate camera poses and a predefined train-test split.
Another popular choice is LLFF~\cite{mildenhall2019llff}, a real-world dataset containing $24$ forward-facing scenes. 
Common categories include $8$ representative scenes: Room, Fern, Leaves, Fortress, Orchids, Flower, T-Rex, and Horns.
The standard protocol holds out one view out of every eight for testing, while the remaining views are used for training.

Several real-world datasets highlight challenges from unbounded indoor and outdoor environments. For example, MipNeRF 360~\cite{Barron2022MipNeRF360} provides $360^\circ$ scene captures with rich high-frequency details, and ETH3D~\cite{schops2017eth3d_multi} offers high-resolution images for large scenes but with sparse spatial coverage. TanksAndTemples~\cite{Knapitsch2017tnt} captures large indoor and outdoor scenes that inevitably contain minor moving objects and variations in exposure.
Phototourism~\cite{jin2021phototourism} is used to evaluate performance under complex in-the-wild conditions, covering variations in lighting and the presence of transient objects over time.

Beyond these, additional datasets are designed to target specific real-world challenges, providing tailored benchmarks for evaluating an approach’s effectiveness.
Shiny Blender~\cite{ref-nerf} and the Shiny dataset~\cite{wizadwongsa2021shiny} specifically focus on the challenges posed by reflective surfaces.
Deblur-NeRF dataset~\cite{ma2022deblurnerf} highlights the low-quality training views with motion blur and camera focus blur.
In the context of dynamic changes, D-NeRF~\cite{d-nerf} and DyNeRF~\cite{li2022neural} datasets feature dynamic objects with slow motions, Nerfies dataset~\cite{Park2021Nerfies} captures casual free-viewpoint selfies, while HyperNeRF dataset~\cite{park2021hypernerf} focuses on the challenges associated with topological changes, e.g., mouth opening.

The novel view synthesis performance is commonly evaluated through three full-reference quality metrics:
\begin{itemize}
    \item \gls{psnr}, which is a distortion metric comparing the mean squared error between the rendered and ground-truth pixels in the $log$ space,
    \item \gls{ssim}~\cite{wang2004ssim}, which is a perceptual metric comparing the similarity of structural details, brightness, and contrast between the rendered and ground-truth image, and
    \item \gls{lpips}~\cite{zhang2018lpips}, which is a perceptual metric comparing the difference of derived features from a deep-learning-based model.
\end{itemize}

\paragraph{3D Surface Reconstruction.}

Three datasets are commonly used for evaluating the 3D surface reconstruction improvement: two object-centric datasets --- DTU~\cite{aanaes2016dtu} and BlendedMVS~\cite{yao2020blendedmvs} --- and TanksAndTemples~\cite{Knapitsch2017tnt} for scenes with a larger scale.
BlendedMVS originally contains $113$ synthetic scenes covering architectures, sculptures, and small objects; within \gls{nerf}-based 3D surface reconstruction techniques, a subset of low-resolution scenes is commonly used, including Dog, Stone, Clock, Sculpture, Bear, Durian, Jade, and Man.
DTU has real-world object scans of different categories and contains various appearances and geometries, among which $15$ scans are commonly used.
It offers RGB images associated with well-calibrated camera poses, 3D point clouds from a structured-light scanner, and observability masks for benchmarking.

During evaluation, the surface (in the form of meshes) is extracted from implicit scene representations, and then a set of points is sampled from the predicted surface.
Typical evaluation metrics used to compare the accuracy of reconstructed surfaces include:
\begin{itemize}
    \item Chamfer distance, quantifying geometric dissimilarity by measuring the average nearest-point distance between predicted and ground-truth sampled point clouds,
    \item F1 score, measuring the percentage of correctly reconstructed points within a threshold distance of the ground truth, and
    \item normal consistency, comparing surface normals at corresponding sampled points to assess the fidelity of local geometric details.
\end{itemize}

\begin{table}[!tb]\centering
\caption{Representative datasets and leading methods for the novel view synthesis and 3D surface reconstruction tasks.}\label{tab:dataset_sota}
\scriptsize
\renewcommand{\arraystretch}{1.5}
\adjustbox{max width=\textwidth}{
\begin{tabular}{l|rrrr}\toprule
Task Category& Scene Content &Representative Datasets &Representative Leading Methods \\\midrule
\multirow{3}{*}{Novel View Synthesis} &Static  &NeRF Synthetic~\cite{Mildenhall2020NeRF}, LLFF-Real~\cite{mildenhall2019llff} &Instant NGP~\cite{instant-ngp}, Mip-NeRF~\cite{Barron2021MipNeRF}, Plenoxels~\cite{Jun2022HDRPlenoxels} \\
&Unbounded &Mip-NeRF 360~\cite{Barron2022MipNeRF360}, Tanks And Temples~\cite{Knapitsch2017tnt} &Zip-NeRF~\cite{Barron2023ZipNeRF}, Mip-NeRF 360~\cite{Barron2022MipNeRF360}, SMERF~\cite{duckworth2024smerf} \\
&Dynamic  &D-NeRF~\cite{d-nerf}, DyNeRF~\cite{li2022neural}, HyperNeRF~\cite{park2021hypernerf} &K-Planes~\cite{Fridovich-Keil2023K-Planes:Appearance}, TiNeuVox~\cite{fang2022tineuvox}, HyperNeRF~\cite{park2021hypernerf} \\ \midrule
\multirow{2}{*}{3D Surface Reconstruction} &Small-scale objects  &DTU~\cite{aanaes2016dtu}, BlendedMVS~\cite{yao2020blendedmvs} &MonoSDF~\cite{monosdf}, Neuralangelo~\cite{neuralangelo} \\
&Large-scale objects &Tanks And Temples~\cite{Knapitsch2017tnt} &Neuralangelo~\cite{neuralangelo} \\
\bottomrule
\end{tabular}
}
\end{table}

\paragraph{Towards Real-world Application.}

Large-scale datasets such as Objaverse~\cite{deitke2023objaverse}, Objaverse-XL~\cite{deitke2023objaversexl}, and CO3D~\cite{reizenstein2021co3d} provide extensive collections of multi-view 3D assets, boosting research for 3D generation and \glspl{nerf}' generalizability to unseen views and scenes.
Indoor datasets like ScanNet++~\cite{yeshwanth2023scannet++} and ARKitScenes~\cite{arkitscenes} offer both 3D ground truth and semantic annotations, supporting research in indoor scene reconstruction, 3D semantic segmentation, as well as robotics.
Large-scale urban scene datasets, including Kitti-360~\cite{liao2022kitti360}, Mill 19~\cite{Turki2022Mega-NeRF}, and Waymo Block-NeRF dataset~\cite{Tancik2022BlockNeRF}, captured either from UAVs or moving vehicles, cover a large area of city- or site-scale environments and serve as valuable benchmarks for large-scale reconstruction and autonomous driving applications.
In addition, realistic datasets with ultra-high-resolution imagery ($4$K/$8$K), such as Eyeful Tower~\cite{xu2023VRNeRF}, have further advanced the development of AR/VR applications.

Tele-health remains one of the most essential application areas for \gls{nerf}-based techniques.
The recent release of diverse medical-related datasets has greatly benefited progress in the reconstruction of medical-related data.
For instance, the MedNeRF dataset~\cite{corona2022mednerf} contributes $20$ CT chest scans and $5$ CT knee scans; 
the EndoNeRF dataset~\cite{EndoNeRF} focuses on endoscopic surgical scenes with tissue deformations; and the SALVE dataset~\cite{chierchia2025salve} offers three scans of 3D wound data associated with ultra-high-resolution $4$K images, supporting development in wound care.

\section{A Conceptual and Decision-oriented Synthesis of Key Methods}
\label{sec1:taxonomy}
\begin{table}[!htp]\centering\small
\caption{A decision-oriented comparison of representative methods.}\label{tab:method_taxonomy}
\scriptsize
\adjustbox{max width=\textwidth}{ 
\begin{tabular}{p{0.12\textwidth}p{0.2\textwidth}p{0.2\textwidth}p{0.2\textwidth}p{0.12\textwidth}p{0.14\textwidth}p{0.15\textwidth}}\toprule
\makecell[l]{\textbf{Method}}  &\makecell[l]{\textbf{Description}} &\makecell[l]{\textbf{Main strengths}} &\makecell[l]{\textbf{Main limitations}} &\makecell[l]{\textbf{Representative}\\\textbf{works}} &\makecell[l]{\textbf{Deployment}\\\textbf{capability}} &\makecell[l]{\textbf{Suitable}\\\textbf{use cases}} \\\midrule
MLP-based &Implicit; use differential volumetric rendering &High reconstruction quality &Slow optimization and rendering; scene specific optimization; sensitive to data density and quality &\cite{Mildenhall2020NeRF} &Ground-laying work &Scene reconstruction in a controlled setting \\ \midrule
Anti-aliased &Implicit or hybrid; use cone tracing to process multi-scale information &High reconstruction quality; handle aliasing artifacts &Slow optimization and rendering &\cite{Barron2021MipNeRF, Barron2022MipNeRF360, Barron2023ZipNeRF} &Rendering fidelity &Scene reconstruction with varying camera resolution or distance \\ \midrule
Surface-constrained &Learn implicit function (e.g., SDF) and apply explicit surface constraints &High geometric accuracy and generate smooth surface &Limited capability to model thin structure and view-dependent effects; slow optimization and rendering &\cite{idr, neus, volsdf, neuralangelo} &Geometric accuracy &Surface reconstruction \\ \midrule
Geometric-regularized &Regularize geometric learning via multi-view consistency checks or the use of external geometric cues (e.g., depths and normals) &Improved geometric accuracy and multi-view consistency &Limited efficiency; reliance on high-quality geometric cues &\cite{monosdf, geo-neus} &Geometric accuracy &Surface reconstruction \\ \midrule
Factorized &Hybrid; factorize feature encodings into low-rank tensors or planes &Fast optimization and rendering; compact; easily extend to 4D scene representations &Trade-offs between resolution and memory requirements; difficult to represent unbounded scenes &\cite{Chen2022TensoRF, Fridovich-Keil2023K-Planes:Appearance} &Computational performance &Efficient static or dynamic scene reconstruction \\ \midrule
Grid/hash-grid-based &Hybrid or explicit; Use voxel-grids/hash-grids to store features or cache predicted colors and densities &Fast optimization and rendering &Limited fidelity for scenes with scale changes; trade-offs between resolution and memory requirements &\cite{instant-ngp, Yu2021PlenOctrees, Jun2022HDRPlenoxels, Sun2021DVGO} &Computational performance &Efficient static scene reconstruction \\ \midrule
Point-based &Hybrid; Store features into points &Fast optimization, improved geometric accuracy &Limited fidelity for complex view-dependent effects; rely on good point initialization; rendering efficiency can be further improved &\cite{Xu2022PointNeRF} &Computational performance &Efficient static scene reconstruction \\ \midrule
3DGS-based &Explicit; use 3D Gaussians and rasterization-based methods to accelerate rendering &Real-time rendering; generally high rendering quality; easily operate on explicit strutures for 3D interaction &Limited fidelity for fine structures and view-dependent appearances; trade-offs between resolution and memory consumption; rely on good initialization point clouds &\cite{kerbl23gaussiansplatting} &Computational performance &Real-time rendering, interactive visualization \\ \midrule
Appearance and transient modeling &Encode the view-dependent appearance and transient objects independently &Handle light changes, appearance variations, and transient objects &Increased model complexity; slow optimization &\cite{nerf-w, Tancik2022BlockNeRF} &Data robustness &Scene reconstruction from unstructured and uncontrolled Internet images, photo-tourism \\ \midrule
Generative &Use generative priors such as GAN or diffusion &Strong capability to predict unseen data &Hallucinated prediction, high computational cost &\cite{niemeyer2020giraffe, chan2022eg3d, liu2023zero123} &Data robustness &Scene reconstruction from sparse views, 3D generation \\ \midrule
Physics-aware modeling &Integrate physical models (e.g., BRDF, imaging process) into image formation process &Improve physical realism and robustness in challenging environments &Increased model complexity; tailored to one domain &\cite{Mildenhall2021RawNeRF, Huang2021HDRNeRF, ref-nerf, levy2023seathru} &Data robustness &Challenging-light reconstruction, non-Lambertian rendering, relighting, underwater robotics \\ \midrule
Pose optimization &Optimize camera parameters and scene reconstruction jointly &Robustness to inaccurate camera poses &Slow optimization; sensitive to local minima &\cite{barf, camp} &Pose robustness &SLAM, scene reconstruction from unstructured image collections \\ \midrule
SLAM-based &Estimate camera trajectory while simultaneously building a scene map &Supports online spatial reasoning &Real-time claims depend on hardware, scene scale, sensor modality, and benchmark setting; robustness remains challenging &\cite{sucar2021imap, zhu2022nice} &Pose robustness &SLAM, localization, incremental mapping \\ \midrule
Space-time modeling &Model the scene flow, object deformation, and non-rigid motion in radiance fields &Simulate non-rigid motion and maintain temporal consistency &Hard to optimize; high training and rendering cost &\cite{d-nerf, Park2021Nerfies, li2020nsff, park2021hypernerf} &Dynamic scene modeling &Dynamic scene reconstruction, human reconstruction \\ \midrule
Spatial decomposition &Split a scene into multiple parts and train NeRF separately &Parallel-training of a large scene &Limited rendering efficiency; limited capability to handle local transient objects &\cite{Turki2022Mega-NeRF, Tancik2022BlockNeRF} &Scalability &Large-scale scene reconstruction \\ \midrule
Generalizable &Image- or feature-conditioned radiance fields &Generalizable to new scene reconstruction without the need to re-train or finetune the network &Limited reconstruction quality on fine-grained details of unseen data &\cite{wang21ibrnet, yu2021pixelnerf, chen2021mvsnerf, hong2024lrm} &Generalizability &Scene reconstruction from sparse views, feed-forward 3D reconstruction \\
\bottomrule
\end{tabular}
}
\end{table}

We summarize the comparisons among representative \gls{nerf}'s advancements and key alternative scene representations in Table~\ref{tab:method_taxonomy}.
Specifically, we organize these methods according to their core mechanisms and compare their main strengths, limitations, and representative application scenarios. 
We further evaluate each method family with respect to practical deployment capabilities, including rendering fidelity, geometric accuracy, computational performance, robustness to imperfect data(e.g., sparse views and unstructured image collections) and inaccurate/unknown camera poses, support for dynamic scene modeling, scalability to large-scale environments, and scene generalizability. 
This comparison provides a decision-oriented reference that aims to help readers understand when different method families are appropriate and what trade-offs they involve under practical constraints such as sparse observations, inaccurate camera poses, degraded imaging conditions, dynamic scenes, large-scale environments, and real-time deployment requirements.

\section{Discussion and Open Challenges}
\label{sec:open_questions}

This survey reviews key \gls{nerf}-related advancements towards higher accuracy and efficiency for representing 3D scenes, and their applicability to real-world problems. 
We covered strategies for enhancing rendering quality and efficiency, analyzed solutions to real-world challenges, and explored applications including reconstructions, 3D generation and editing, recognition, and robotics. 
Finally, we compiled a list of frequently used datasets and useful tools for \gls{nerf}-related research. 
This section takes a step back and explores open challenges for advancing \glspl{nerf} in the future.

Current \gls{nerf} enhancements yield impressive results with images up to 1K resolution.
However, when handling higher resolutions such as 4K, the rendering time is multiplied by an order of magnitude compared to 1K images, limiting widespread adoption~\cite{li2023uhdnerf}.
The rise of inexpensive mobile devices, capable of acquiring high-resolution images but limited in computing resources, highlights the need to optimize \glspl{nerf} towards higher efficiency and compactness with minimal GPU requirements.
One approach is to incorporate explicit representations such as 3D Gaussian Splatting~\cite{kerbl23gaussiansplatting} as they can achieve real-time renderings.
This could involve creating a hybrid representation or transcoding a \gls{nerf} into alternative explicit representations.
Additionally, data compression techniques like quantization might be helpful to improve the model's compactness~\cite{Hedman2021SNeRG}.

As \glspl{nerf}' performance continues to improve, it is crucial to revisit the choice of testing sets~\cite{xiao2024nerfdirector} and evaluation metrics~\cite{qu2024nerfnqa}. Current evaluations heavily rely on reference 2D images, making them sensitive to testing camera distributions, testing view quality, and scene occlusions.


In-the-wild scenes pose multiple challenges, including heterogeneous image quality, limited observations, diverse dynamics (e.g., transient objects and complex lighting/weather variations), and large-scale scenes. 
Developing methods that can efficiently and accurately model such scenes while robustly handling distractors remains an active area of research~\cite{sabour2023robustnerf}.

Recent research~\cite{Zhou2023NeRFLix, wu2024rafe} applies recent blind image restoration techniques to improve the quality of \gls{nerf} trained on degraded inputs, and we expect more effective and efficient restoration systems to emerge. 
Several few-shot \glspl{nerf} utilize learned priors from pre-trained deep models (e.g., monocular depth priors) to guide training under limited views; however, ensuring metric accuracy and maintaining multi-view consistency in priors remains an open challenge~\cite{Wang2023SparseNeRF}.
In addition, limited views make \gls{sfm}-based camera pose estimation underconstrained and prone to inaccuracy.
Two promising directions include: i) developing online refinement algorithms for feature correspondences and camera parameters~\cite{camp, sparf}, and ii) leveraging feed-forward 3D reconstruction methods, e.g., DUSt3R~\cite{wang2024dust3r} and MASt3R~\cite{leroy2024mast3r}.

Accurately and efficiently modeling realistic non-rigid deformations remains a long-standing and important challenge in dynamic \glspl{nerf}, where hybrid representations may help~\cite{Fridovich-Keil2023K-Planes:Appearance}.
In-the-wild dynamic scenes involve not only local deformations (e.g.,  articulation/motion and topological changes) but also global variations (e.g., illumination/weather changes).
Capturing both scales efficiently remains challenging~\cite{ramasinghe2024blirf}.
Physics-aware methods prove effective in modeling complex outdoor scene lighting~\cite{nerf-osr} and soft-body deformations~\cite{qiao2022neuphysics}.
Integrating richer physics simulation priors offers a promising path toward physically grounded dynamic \gls{nerf} models.

Despite significant advances in large-scale scene modeling, challenges of memory scalability and computational efficiency remain open.
Achieving streamable performance in large-scale environments remains an active research direction for interactive VR/AR~\cite{duckworth2024smerf}, e.g., enabling exploration of multi-room spaces in real-estate applications.

In addition, improving uncertainty quantification and generalizability of \glspl{nerf} is essential for practical real-world use.
Open questions in uncertainty quantification include: 
1) What constitutes well-quantified uncertainty for  \glspl{nerf}? 
2) Can we efficiently quantify both epistemic and aleatoric uncertainty?
For generalizability, regardless of achieving impressive renderings of unseen scenes, challenges remain with lengthy training time and insufficient details in unobserved areas.

The field of 3D vision has evidenced the broad adoption of \glspl{nerf}, transforming 3D surface reconstruction across various data types e.g., medical imaging\cite{cunerf}.
Real-world applications, such as robotics, require the handling of diverse signals. Recent research demonstrates the potential of using multi-modal inputs---such as semantic labels~\cite{zhi2021place}, event streams~\cite{Qi2023E2NeRF}, multispectral data~\cite{li2024specnerf}, and audio~\cite{guo2021adnerf}---to inform and enhance 3D scene learning.
Although in its early stages, this research direction presents a promising yet underexplored area: how to efficiently leverage priors from diverse signals for 3D scene learning.
Another exciting challenge lies in learning from modalities beyond traditional visual data, \textit{e.g.,} developing \gls{nerf}-like frameworks from LiDAR sensors~\cite{tao2024lidarnerf}.

Beyond reconstruction tasks, \gls{nerf} increasingly impacts traditional computer vision tasks, such as 3D semantic segmentation.
Learning 3D scenes naturally supports self-supervised learning.
It is exciting to see how \glspl{nerf} can seamlessly integrate with and improve various large-scale 3D downstream tasks~\cite{lerf2023}.
Moreover, advancements in large models may further boost the end-to-end learning of \glspl{nerf}, potentially bypassing the traditional camera pose estimation followed by reconstruction steps~\cite{wang2025vggt}. This could lead to direct camera pose estimation and 3D geometry learning with large models like Transformers.

\gls{nerf} has also refashioned 3D generation and editing tasks, though existing \gls{nerf}-based 3D generation techniques are limited to dream-style object generation, often lacking fine-grained realistic details and requiring long optimization times~\cite{poole2023dreamfusion}.
Developing high-fidelity, real-time, faithful, and controllable 3D generation techniques would benefit AR/VR and creative industries.
Conversely, generative models can be used to improve reconstruction.
There is a trend in using generative priors learned from large-scale image assets to enrich details on unobserved areas and facilitate scene learning, although decreasing their computational overheads presents an ongoing challenge~\cite{liu2023zero123}.
Moreover, it is crucial to consider the ethical and copyright implications of these generative methods.

\section*{Acknowledgments}
We thank Rongkai Ma and Ethan Goan for their valuable feedback on our manuscript. We also thank Ines Vati for providing sources of several figures used in our manuscript.
This work was supported in part by the Australian Research Council (ARC) Discovery research grant DP250103634 and 
the CSIRO AI4M Research Program.

\bibliographystyle{ACM-Reference-Format}
\bibliography{target}

\end{document}